\documentclass[lettersize,journal]{IEEEtran}
\usepackage{amsmath,amsfonts}
\usepackage{algorithmic}
\usepackage{algorithm}
\usepackage{array}
\usepackage[caption=false,font=normalsize,labelfont=sf,textfont=sf]{subfig}
\usepackage{textcomp}
\usepackage{stfloats}
\usepackage{url}
\usepackage{verbatim}
\usepackage{graphicx}
\usepackage{cite}
\usepackage{color}
\begin{document}
%
\title{Asynchronous Curriculum Experience Replay: A Deep Reinforcement Learning Approach for UAV Autonomous Motion Control in Unknown Dynamic Environments}


\author{\IEEEauthorblockN{Zijian Hu\IEEEauthorrefmark{1},
Xiaoguang Gao\IEEEauthorrefmark{1},~\IEEEmembership{Member,~IEEE},
Kaifang Wan\IEEEauthorrefmark{1},
Qianglong Wang\IEEEauthorrefmark{1}, and
Yiwei Zhai\IEEEauthorrefmark{1}}

\IEEEauthorblockA{\IEEEauthorrefmark{1}School of Electronics and Information,
Northwestern Polytechnical University, Xi'an, Shaanxi 710129 China}
\thanks{Manuscript received December 1, 2012; revised August 26, 2015. 
Corresponding author: Kaifang Wan (email: wankaifang@nwpu.edu.cn).}}

\markboth{Journal of \LaTeX\ Class Files,~Vol.~14, No.~8, August~2015}%
{Shell \MakeLowercase{\textit{et al.}}: Bare Demo of IEEEtran.cls for IEEE Transactions on Magnetics Journals}
%




\maketitle

\IEEEdisplaynontitleabstractindextext{%
\begin{abstract}
\emph{Abstract}\----Unmanned aerial vehicles (UAVs) have been widely used in military warfare. In this paper, we formulate the autonomous motion control (AMC) problem as a Markov decision process (MDP) and propose an advanced deep reinforcement learning (DRL) method that allows UAVs to execute complex tasks in large-scale dynamic three-dimensional (3D) environments. To overcome the limitations of the prioritized experience replay (PER) algorithm and improve performance, the proposed asynchronous curriculum experience replay (ACER) uses multithreads to asynchronously update the priorities, assigns the true priorities and applies a temporary experience pool to make available experiences of higher quality for learning. A first-in-useless-out (FIUO) experience pool is also introduced to ensure the higher use value of the stored experiences. In addition, combined with curriculum learning (CL), a more reasonable training paradigm of sampling experiences from simple to difficult is designed for training UAVs. By training in a complex unknown environment constructed based on the parameters of a real UAV, the proposed ACER improves the convergence speed by 24.66\% and the convergence result by 5.59\% compared to the state-of-the-art twin delayed deep deterministic policy gradient (TD3) algorithm. The testing experiments carried out in environments with different complexities demonstrate the strong robustness and generalization ability of the ACER agent.
\end{abstract}
	
}


%
\IEEEpeerreviewmaketitle

\section{Introduction}
\begin{figure}[htb]
\centering
\includegraphics[width=3.5in]{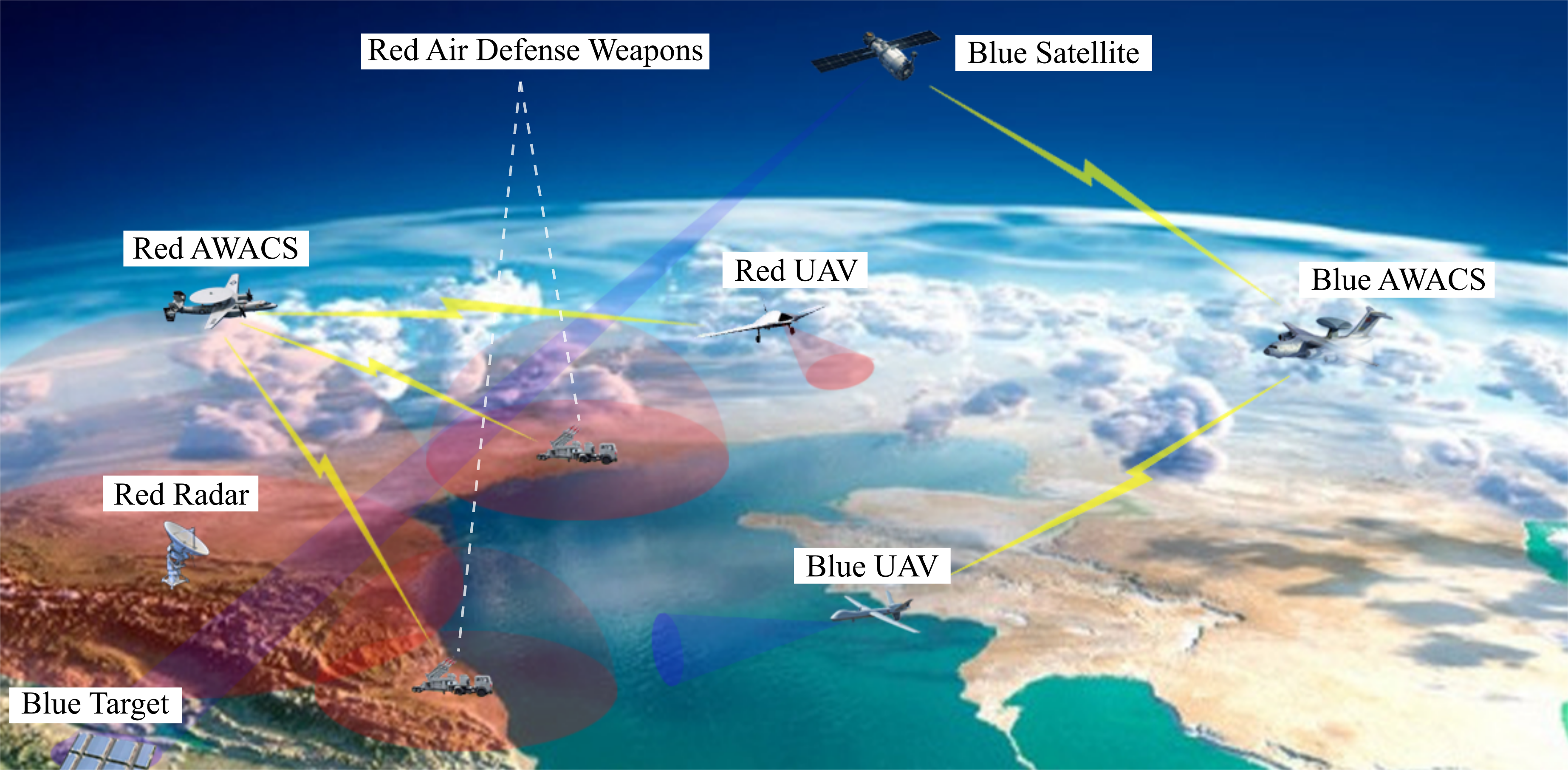}
\caption{Operational scenario of the blue side ground attack mission. The objective of the blue side  is to destroy the target building behind the red side at the lowest cost, while the objective of the red side is to protect the building. During the whole operation, the red air defense facilities are in motion.}
\label{fig1}
\end{figure}

%
%
%
%
\IEEEPARstart{W}{ith} the rapid development of unmanned aerial vehicle (UAV) technology, UAVs have been widely used in military wars in recent years. The characteristics of UAVs, such as low cost, strong survivability and high operational efficiency, make them perform well in executing tasks such as intelligence, surveillance, and reconnaissance \cite{1,2}, electronic countermeasures \cite{3,4}, and ground attacks \cite{5,6}. To accomplish these tasks successfully, UAVs usually need to achieve autonomous motion control (AMC) in complex and changeable unknown environments. Taking the ground attack as an example (Fig. \ref{fig1}), the blue UAV needs to approach the target as close as possible with the assistance of the satellite and the airborne warning and control system (AWACS) and avoid hitting the mountain or being detected by the red defense facilities (red UAV, red AWACS, red radar, and red air defense weapons) during the flight.

The results of UAV AMC will directly affect the results of the combat mission, so a long-term goal of UAV applications is to develop a technique that can enable UAVs to fly safely and accurately without human intervention. As a popular research topic, UAV AMC has aroused widespread interest, and a series of methods have been proposed to address it. These methods can be generally classified into two groups: nonlearning-based and learning-based \cite{7}. Many studies \cite{8,9} used nonlearning-based methods such as the A* algorithm for UAV route planning and performed well when information for the entire environment is known. To adapt to unknown or partially known environments, another class of nonlearning-based methods \cite{10,11} resorts to simultaneous localization and mapping to control UAVs. Since such methods need to model the environment based on observational information, once the environment changes, modeling will lead to unaffordable computational costs \cite{12}. Therefore, it is necessary to find more efficient methods to control UAVs in highly dynamic unknown environments.

Reinforcement learning (RL), an online learning-based method, has been introduced to address the AMC problem. Li \cite{13} presented a novel route planning algorithm based on Q-learning \cite{14}, which improves the convergence speed by using prior knowledge to guide the UAV in selections. Hung \cite{15} applied Q-learning to flocking so that UAVs can learn how to flock in a simulated nonstationary stochastic environment. As the complexity of the environment continues to increase, deep reinforcement learning (DRL), a new area of intense interest that combines the perceived capabilities of deep learning and the decision-making capabilities of RL, has been proven to be a more effective approach to solve the problem. Kersandt \cite{16} applied deep Q-learning (DQN) \cite{17}, double DQN (DDQN) \cite{18} and dueling DQN \cite{19} in the same UAV control mission and compared the results of these algorithms. Singla \cite{20} proposed a deep recurrent Q-network with temporal attention that enables a UAV to autonomously avoid obstacles in unknown environments. By leveraging DDQN with multistep learning, Zeng \cite{21} introduced a UAV navigation algorithm that uses signal measurement to directly train the action-value function of the navigation policy. However, the action space of UAVs in the battlefield environment is continuous, which requires policy-based DRL methods to control the UAV. Based on the deep deterministic policy gradient (DDPG) \cite{22} algorithm, Ding \cite{23} proposed an algorithm named energy-efficient fair communication with trajectory designs and band allocation to implement UAV control in auxiliary communication tasks. For the UAV ground target tracking task, Li \cite{24} introduced an improved DDPG algorithm that uses long short-term memory networks to approximate the state of environments and improves the approximation accuracy and the efficiency of data utilization.

New challenges arise when UAVs perform AMC in realistic large-scale environments: the limited number of experiences. In military warfare, the preparation time for UAVs to perform each combat mission is usually limited. Limited training time results in limited experience. Therefore, it is necessary to design an algorithm to use limited experience to train the best UAV. To solve this problem, usually one of two approaches is chosen: 1) Create more reliable experiences \cite{25,26}. Ensuring the accuracy and reliability of the generated experiences in this approach, however, is difficult. 2) Use limited experiences more efficiently. This method is the focus of this paper, and it is also a research hotspot of DRL: Experience Replay (ER).

To mitigate the pressure on training caused by the lack of reliable experiences, Schaul \cite{27} first introduced a prioritized experience replay (PER) approach that greatly improves the convergence speed of DRL algorithms by replaying important experiences more frequently. Although PER has shown advantages in many scenarios, it still has the following limitations: 1) Updating the temporal difference error (TD error) too slowly that affects the value of the sampled experiences. 2) Using the clip operation weakens the difference between experiences. 3) Maximum priority cannot ensure the priorities of the newest experiences. 4) Sampling according to the TD error alone may not be the best method. 
Many studies have made efforts to solve these problems: Ren \cite{28} proposed a deep curriculum reinforcement learning (DCRL) method that combines DRL with curriculum learning (CL) \cite{29} to adaptively select appropriate experiences based on the complexity of each experience. Our previous work \cite{12} considered the relevance between the state and experiences and presented a relevant experience learning (REL) method to use more indicators to replay experiences. These novel methods only solve the fourth limitation of PER to some extent, while this paper attempts to fundamentally overcome all of the PER limitations and makes the following contributions:

1) The AMC problem of UAV in unknown complex environments is modeled as a Markov decision process (MDP). The state space, action space, and reward function are well designed for the MDP so that the problem can be addressed by different DRL algorithms.

2) Based on CL, a DRL framework for controlling UAVs in large-scale environments is developed for the first time. Under this framework, algorithms can change the training paradigm of agent to improve the robustness of the DRL algorithms.

3) An efficient ER algorithm, the asynchronous curriculum experience replay (ACER), is proposed to overcome all the limitations of PER. ACER uses a subthread to asynchronously update the priorities, assigns the true priorities and makes full use of the new experiences and the stored experiences to improve the convergence speed.

4) By combining with the state-of-the-art twin delayed deep deterministic policy gradient (TD3) algorithm \cite{30}, many experiments were conducted on the environments described in \cite{12}. The training experiments show that ACER improves the convergence speed by 24.66\% and the convergence result by 5.59\% compared with vanilla TD3. Testing experiments in environments with different complexities prove the better robustness and generality of the ACER algorithm. In addition, the motion trajectories of the ACER agent show the practicality of the proposed algorithm for future use on real UAVs.

The remainder of this paper is organized as follows. In Section II, some background knowledge of TD3 and PER is introduced. Section III formulates the UAV AMC problem as MDP. The proposed algorithm is presented in Section IV. In Section V, training and testing experiments are presented, and the experimental results are discussed. Section VI concludes this paper and envisages some future work.

\section{Background}
In this section we first give a brief introduction of a state-of-the-art off-policy DRL algorithm, followed by a novel ER algorithm.

\subsection{Actor-Critic, DDPG, and TD3}
Value-based methods such as DQN, have performed well in addressing problems with discrete action space. In continuous spaces, finding a greedy policy requires optimization at every timestep, but this optimization is too slow to be practical with large, unconstrained function approximators and is nontrivial \cite{22}. Unlike value-based methods, policy-based methods approximate the policy directly so that they can perform well in searching action in continuous spaces.

Actor-critic \cite{31}, a category of policy-based methods addressing MDPs\cite{32,33}, seeks to learn the optimal policy $\boldsymbol{a}\!\sim\! \pi_{\boldsymbol{\theta}}\!(\boldsymbol{s})$ by applying a stochastic policy gradient in the parameter space: 

\begin{equation}
\nabla_{\boldsymbol{\theta}} J\left(\pi_{\boldsymbol{\theta}}\right)=\mathbb{E}_{\boldsymbol{s} \sim \rho^{\pi}, \boldsymbol{a} \sim \pi_{\boldsymbol{\theta}}}\left[\nabla_{\boldsymbol{\theta}} \log \pi_{\boldsymbol{\theta}}(\boldsymbol{s}, \boldsymbol{a}) Q^{\pi}(\boldsymbol{s}, \boldsymbol{a})\right],
\label{equ1}
\end{equation}
where $\rho^{\pi}$ is the space of the sampled states and $\pi_{\boldsymbol{\theta}}$ is the action space. The stochastic policy gradient method needs many experiences to sample the whole action space, while the deterministic policy gradient methods select the action $\boldsymbol{a}=\mu_{\boldsymbol{\theta}}(\boldsymbol{s})$ with the highest probability at every state, which reduces the amount of experience sampling and improves the efficiency of the algorithm. The gradient of the deterministic policy methods is as follows:

\begin{equation}
\nabla_{\boldsymbol{\theta}} J\left(\mu_{\boldsymbol{\theta}}\right)=\mathbb{E}_{\boldsymbol{s} \sim \rho^{\mu}}\left[\left.\nabla_{\boldsymbol{\theta}} \mu_{\boldsymbol{\theta}}(\boldsymbol{s}) \nabla_{\boldsymbol{a}} Q^{\mu}(\boldsymbol{s}, \boldsymbol{a})\right|_{\boldsymbol{a}=\mu_{\boldsymbol{\theta}}(\boldsymbol{s})}\right],
\label{equ2}
\end{equation}
where $\rho^{\mu}$ is the space of the sampled states. To optimize the training process, Lillicrap \cite{22} proposed an actor-critic method DDPG, which shows good results in solving MDPs with continuous action spaces. Derived from the techniques of DQN, both the actor and critic networks of DDPG contain two artificial neural networks with the same structure, called eval net and target net. The parameters of eval nets are updated more frequently than those of target nets to make the algorithm more stable \cite{17}. In addition, DDPG adds independent noise $\mathcal{N}_{t}$ to increase the randomness of the agent's exploration:

\begin{equation}
\boldsymbol{a}_{t}=\mu\left(\boldsymbol{s}_{t} \mid \boldsymbol{\theta}^{\mu}\right)+\mathcal{N}_{t}.
\label{equ3}
\end{equation}

The actor network uses policy gradient $\nabla_{\boldsymbol{\theta}^{\mu}} J$ to approximate the parameters of eval net while the critic network updates its eval net by minimizing the loss function $L\left(\boldsymbol{\theta}^{Q}\right)$:

\begin{equation}
\nabla_{\boldsymbol{\theta}^{\mu}}\!J\!\approx\!\frac{1}{N} \sum_{i}\!\nabla_{\boldsymbol{a}}Q\!\left(\!\boldsymbol{s}, \boldsymbol{a}\!\mid\!\boldsymbol{\theta}^{Q}\right)\!\mid _{\boldsymbol{s}=\boldsymbol{s}_{i}, \boldsymbol{a}=\mu\left(\boldsymbol{s}_{i}\!\right)\!} \!\nabla_{\boldsymbol{\theta}^{\mu}} \mu\!\left(\boldsymbol{s} \!\mid\!\boldsymbol{\theta}^{\mu}\right)\!\mid\! {\boldsymbol{s}_{i}},
\label{equ4}
\end{equation}
\begin{equation}
L\left(\boldsymbol{\theta}^{Q}\right)=\frac{1}{N} \sum_{i}\left(y_{i}-Q\left(\boldsymbol{s}_{i}, \boldsymbol{a}_{i} \mid \boldsymbol{\theta}^{Q}\right)\right)^{2},
\label{equ5}
\end{equation}
\begin{equation}
y_{i}=r\left(\boldsymbol{s}_{i}, \boldsymbol{a}_{i}\right)+\gamma Q^{\prime}\left(\boldsymbol{s}_{i+1}, \mu^{\prime}\left(\boldsymbol{s}_{i+1} \mid \boldsymbol{\theta}^{\mu^{\prime}}\right) \mid \boldsymbol{\theta}^{Q^{\prime}}\right),
\label{equ6}
\end{equation}
where $\boldsymbol{\theta}^{\mu}$, $\boldsymbol{\theta}^{\mu^{\prime}}$, $\boldsymbol{\theta}^{Q}$, and $\boldsymbol{\theta}^{Q^{\prime}}$ represent the parameters of the eval net in the actor network, the target net in the actor network, the eval net in the critic network, and the target net in the critic network, respectively, and $N$ is the number of sampled experiences. 

\begin{figure}[h]
	\centering
	\includegraphics[width=3.5in]{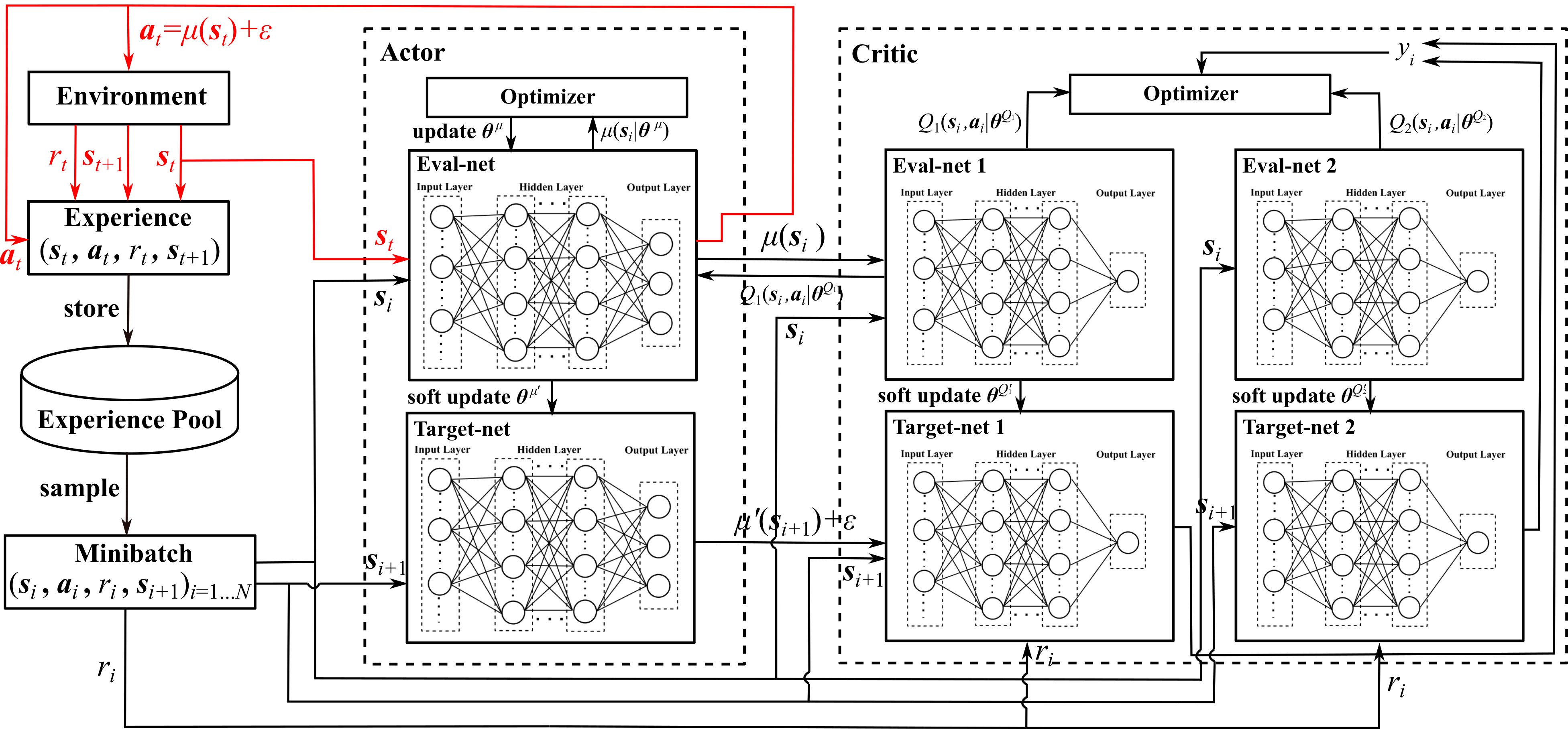}
	\caption{The framework of the TD3 algorithm. The red arrows show the interaction process between the agent and the environment, and the black arrows present the process of a minibatch being sampled from the experience pool and learned to update network parameters.}
	\label{fig2}
\end{figure}

Based on the DDPG algorithm, Fujimoto \cite{30} introduced a novel policy-based method named TD3, which has been proven to be one of the state-of-the-art DRL algorithms. As shown in Fig. \ref{fig2}, the TD3 algorithm also uses an experience pool for storing and replaying old experiences. The actor network is used to determine the probability of the agent choosing an action while the critic networks are used to evaluate the action selected by the agent based on the environment state.

TD3 makes the following enhancements to improve the performance: 1) Double networks for overestimation: Like the DDQN algorithm, TD3 proposes to use two sets of critic networks to calculate $Q(\boldsymbol{s}, \boldsymbol{a})$ and chooses the smaller set as the target. Therefore,  (\ref{equ5}) and (\ref{equ6}) become:
\begin{equation}
L\left(\boldsymbol{\theta}^{Q_{j}}\right)=\min _{j=1,2} \frac{1}{N} \sum_{i}\left(y_{i}-Q_{j}\left(\boldsymbol{s}_{i}, \boldsymbol{a}_{i} \mid \boldsymbol{\theta}^{Q_{j}}\right)\right)^{2},
\label{equ7}
\end{equation}
\begin{equation}
y_{i}\!=\!r\!\left(\boldsymbol{s}_{i}, \boldsymbol{a}_{i}\right)\!+\!\gamma \min _{j=1,2} Q_{j}^{\prime}\left(\boldsymbol{s}_{i+1}, \mu^{\prime}\left(\boldsymbol{s}_{i+1} \!\mid\! \boldsymbol{\theta}^{\mu^{\prime}}\right) \!\mid\! \boldsymbol{\theta}^{Q_{j}^{\prime}}\right).
\label{equ8}
\end{equation}
2) Actor update delay for stability: Different from the synchronous updates of DDPG, TD3 allows critic networks to update more frequently than actor networks to make the training of actors more stable. 3) Action noise for smooth target policy: TD3 adds action noise $\varepsilon \sim \operatorname{clip}(\mathcal{N}(0, \tilde{\sigma}),-l, l)$ to make an action random in a certain range when calculating $Q$ to make the policy smooth and stable. 

The parameter update method of TD3 is the same soft-updated method as that in the DDPG algorithm:
\begin{equation}
\boldsymbol{\theta}^{\prime} \leftarrow \tau \boldsymbol{\theta}+(1-\tau) \boldsymbol{\theta}^{\prime},
\label{equ9}
\end{equation}
where $\tau \in[0,1]$ determines the update degree of the network parameters. 

\subsection{PER}
ER, a process of storing past sequential experiences and sampling them to reuse for updating policies, was first explored by Lin \cite{34} in 1992 to speed up the training process. More recently, ER has seen widespread adoption, since it has been shown to be instrumental in the breakthrough success of DRL \cite{17}. This new area of intense research has attracted many studies to explore how ER can influence the performance of off-policy DRL algorithms\cite{35,36,37}. ER is always a fixed-size experience pool that holds the most recent experiences collected by the agent. This first-in-first-out (FIFO) buffer brings two advantages: 1) uniform sampling breaks the correlation between experiences to improve the stability of the policy; 2) the large-capacity buffer ensures the possibility of learning from long-term experiences, thereby avoiding 'catastrophic forgetting' \cite{38}.

In the ER of DQN, the uniform sampling policy results in all experiences having the same probability of being sampled, which ignores the importance of each experience. PER \cite{27} differs from this and assigns the priority of the experience $\boldsymbol{e}_{i}$ in the experience pool according to its TD error:
\begin{equation}
	\delta_{i}=r_{i}+\gamma \max _{\boldsymbol{a}_{i+1}} Q^{\prime}\left(\boldsymbol{s}_{i+1}, \boldsymbol{a}_{i+1}\right)-Q(\boldsymbol{s}, \boldsymbol{a}),
	\label{equ10}
\end{equation}	
\begin{equation}
	p_{i}=\left|\delta_{i}\right|+\varepsilon.
	\label{equ11}
\end{equation}

The sampling probability of $\boldsymbol{e}_{i}$ is defined as:
\begin{equation}
	P\left(\boldsymbol{e}_{i}\right)=\frac{p_{i}^{\alpha}}{\sum\limits_{j=1}^{D} p_{j}^{\alpha}},
	\label{equ12}
\end{equation}
where $D$ is the capacity of the experience pool and $\alpha \in[0,1]$ is a constant to control the size of $p_{j}^{\alpha}$. These ensure that the greater the TD error is, the higher the priority of the experience is and the greater the probability is that the experience can be sampled. To reduce the computational complexity of searching for experiences with higher priorities, PER adopts a 'sum-tree' structure to improve search efficiency and ensure the priority and diversity of the sampled experiences at the same time \cite{39}. Following the innovation of PER, Brittain \cite{40} believed that the previous experiences leading to the important experiences with higher TD error should also be assigned higher priorities and proposed the prioritized sequence experience replay algorithm.

In addition, some studies try to improve the PER algorithm from different perspectives. Some of them believed that TD error should not be the only criterion for measuring the importance of experience, and other indicators, such as reward \cite{41,42}, difficulty \cite{28} and the relevance between experiences and current state \cite{12}, should also be considered. Ren \cite{28} first introduced CL into ER and proposed DCRL to use complexity and coverage penalties to control the experiences learned by the agent. These methods improved the convergence speed of PER to a certain extent, but they did not fundamentally overcome the limitations of the PER algorithm.

\section{Problem Formulation}
In this section, the AMC problem is formulated for complex unknown environments. The UAV model is introduced first. Then, the details of the MDP modeling procedure are presented.

\subsection{UAV Model}
UAV is usually equipped with autopilot system to provide low-level flight control, that is, the autopilot system controls the thrust system and wings to generate the required dynamic parameters, so as to realize the control of altitude, forward and vertical speeds, and the pitch attitude, and finally drive the aircraft to fly stably or track the desired path \cite{43}. Because RL algorithm usually provides the waypoints needed by UAV to achieve AMC, this paper only considered the high-level flight control model of UAV and used the load factor $\boldsymbol{n}_{\rm{u}}$ as the input to control the UAV. According to reference \cite{44,45,46}, assuming that $\boldsymbol{N}$ is the sum of all external forces acting on the UAV except gravity, the definition of $\boldsymbol{n}_{\rm{u}}$ is as follows:
\begin{equation}
	\boldsymbol{n}_{\mathrm{u}}=\frac{\boldsymbol{N}}{G},
	\label{equ13}
\end{equation}
where $G$ is the dimensionless value of gravity $\boldsymbol{G}$. The centroid acceleration $\boldsymbol{a}_{\rm{u}}$ of UAV can be expressed as:
\begin{equation}
	\boldsymbol{a}_{\mathrm{u}}=\frac{\boldsymbol{N}+\boldsymbol{G}}{m_{\mathrm{u}}},
	\label{equ14}
\end{equation}
where $m_\mathrm{u}$ is the mass of UAV. And $\boldsymbol{a}_{\mathrm{u}}$ can be represented by $\boldsymbol{n}_{\rm{u}}$:

\begin{equation}
	\boldsymbol{a}_{\mathrm{u}}=\boldsymbol{n}_{\mathrm{u}}\cdot g + \boldsymbol{g}.
	\label{equ15}
\end{equation}
where $g$ is the dimensionless value of gravitational acceleration $\boldsymbol{g}$. Then the velocity $\boldsymbol{v}_{\rm{u}}$ and position $\boldsymbol{p}_{\mathrm{u}}$ of the UAV can be further calculated:
\begin{equation}
\boldsymbol{v}_{\mathrm{u}}=\boldsymbol{v}_{0}+\int_{t} \boldsymbol{a}_{\mathrm{u}} d t,
\label{equ16}
\end{equation}
\begin{equation}
\boldsymbol{p}_{\mathrm{u}}=\boldsymbol{p}_{0}+\int_{t} \boldsymbol{v}_{\mathrm{u}} d t.
\label{equ17}
\end{equation}
where $\boldsymbol{v}_{0}$ and $\boldsymbol{p}_{0}$ is the velocity and position at the previous timestep, respectively.
\begin{figure}[H]
	\centering
	\includegraphics[width=3.5in]{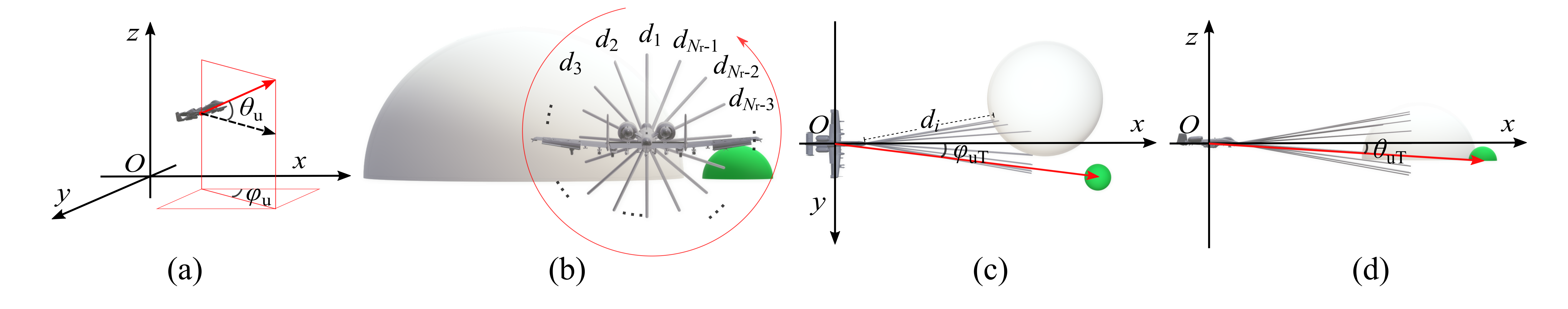}
	\caption{(a). UAV attitude angles calculated in 3D. The pitch angle $\theta_{\mathrm{u}}$ is the angle between the UAV axis and the plane formed by the $x$ and $y$ axes, while the yaw angle $\varphi_{\mathrm{u}}$ is the angle between the UAV axis and the $x$ axis. (b), (c) and (d) are the back view, the top view, and the side view of UAV sensing information from the environment, respectively. The white hemisphere is the obstacle, while the green hemisphere represents the target. $d_{1}, d_{2}, \cdots, d_{N_{\mathrm{r}}}$ denote the distances returned by the airborne radar.  $\varphi_{\mathrm{uT}}$ and $\theta_{\mathrm{uT}}$ denote the yaw and pitch angles between the UAV and the target, respectively.}
	\label{fig3}
\end{figure}

Assuming that the velocity direction of the UAV is always the same as the UAV axis (shown as the red arrow in Fig. \ref{fig3}(a)), the pitch angle $\theta_{\mathrm{u}}$ and yaw angle $\varphi_{\mathrm{u}}$ of the UAV at the current timestep can be calculated according to the velocity direction:

\begin{equation}
\theta_{\mathrm{u}}=\operatorname{atan}\left(\frac{v_{\mathrm{u}z}}{\sqrt{v_{\mathrm{u}x}{ }^{2}+v_{\mathrm{u}y}{ }^{2}}}\right),
\label{equ18}
\end{equation}
\begin{equation}
\varphi_{\mathrm{u}}=\operatorname{atan}\left(\frac{v_{\mathrm{u}y}}{v_{\mathrm{u}x}}\right).
\label{equ19}
\end{equation}

\subsection{MDP Modeling}
An MDP is always represented as a 4 tuple $(\boldsymbol{S}, \boldsymbol{A}, P, R)$: $\boldsymbol{S}$ is a set of all states that the agent can get in the environment; $\boldsymbol{A}$ represents the set of all actions that can be chosen by the agent in the environment; $P$ is the probability of executing an action $\boldsymbol{a}$ from state  $\boldsymbol{s}$ to $\boldsymbol{s}^{\prime}$(where $\boldsymbol{a} \in \boldsymbol{A}$ and $\boldsymbol{s}, \boldsymbol{s}^{\prime} \in \boldsymbol{S}$); $R$ represents the reward of performing action $\boldsymbol{a}$ at state $\boldsymbol{s}$ \cite{32}. The UAV AMC problem is a typical sequential decision-making problem that can be modeled as MDP.
\subsubsection{State Space and Action Space Specification}
For autonomous control, the UAV shall at least be capable of collecting information from three sources, i.e., the information of its own state, the information observed from the environment, and the information of the target.

In a real combat mission, the position, velocity, and attitude of the UAV can be provided by onboard GPS, sensors, and gyroscope devices. In this paper, the state of the UAV is defined as a vector: $\boldsymbol{\xi}_{\mathrm{u}}=\left[p_{\mathrm{u}x}, p_{\mathrm{u}y}, p_{\mathrm{u}z}, \varphi_{\mathrm{u}}, \theta_{\mathrm{u}},\left\|\boldsymbol{v}_{\mathrm{u}}\right\|\right]^{\mathrm{T}}$. For environmental information, we use $N_{\mathrm{r}}$ segments to simulate radar rays of a terrain avoidance radar (TAR) \cite{47}, which is mainly used to ensure the safety of UAVs when flying at low altitudes. As shown in Fig. \ref{fig3} (b) and (c), the detected relative distance $d_{i}$ between the UAV and obstacles can be returned by the $i$-th ray. All the radar returns are grouped together from the environmental information $\boldsymbol{\xi}_{\mathrm{e}}=\left[d_{1}, d_{2}, \cdots, d_{N_{\mathrm{r}}}\right]^{\mathrm{T}}$, which indicates the safety of the state of the UAV. In the actual battlefield environment, the position of the target is usually obtained by satellite and transmitted to AWACS for further translation and transmission to UAVs. Here, a vector $\boldsymbol{\xi}_{\mathrm{T}}=\left[p_{\mathrm{T} x}, p_{\mathrm{T} y}, p_{\mathrm{T} z}\right]^{\mathrm{T}}$ indicates that the target coordinates are supposed to be transmitted to the UAV periodically. Subsequently, $\boldsymbol{\xi}_{\mathrm{u}}$, $\boldsymbol{\xi}_{\mathrm{e}}$, and $\boldsymbol{\xi}_{\mathrm{T}}$ are combined to form the ${N_{\rm{r}}}{\rm{+ 6}}$ dimensional vector state $\boldsymbol{s}$ of the MDP:$\boldsymbol{s}=\left[p_{\mathrm{T}x}\!-\!p_{\mathrm{u}x},p_{\mathrm{T}y}\!-\!p_{\mathrm{u}y},p_{\mathrm{T}z}\!-\!p_{\mathrm{u}z},\varphi_{\mathrm{u}},\theta_{\mathrm{u}},\!\left\|\boldsymbol{v}_{\mathrm{u}}\right\|\!,  d_{1},d_{2},\!\cdots\!,d_{N_{\mathrm{r}}}\!\right]^{\mathrm{T}}$.

As introduced before, the UAV can be controlled by the required load factor $\boldsymbol{n}_{\rm{u}}$ in 3D environments; thus, the action $\boldsymbol{a}$ of the MDP can be designed as $\boldsymbol{a}=\left[n_{\mathrm{u}x}, n_{\mathrm{u}y}, n_{\mathrm{u}z}\right]^{\mathrm{T}}$.

\subsubsection{Reward Design}
Reward acts as the only criterion for evaluating how good the action chosen by the agent at a certain state is. The design of the reward function will have a huge impact on the convergent motion policy of the agent. A reasonable reward function can speed up the convergence of the algorithm, and an unreasonable reward function may cause the algorithm to not converge. When controlling UAV in large-scale environments, due to the initial policy of DRL is randomly generated, algorithms would take an extremely long time to converge if a sparse reward function is designed \cite{7}. In this paper, a nonsparse reward function that incorporates domain knowledge about the AMC problem is introduced. This nonsparse reward function consists of five parts, namely, position reward, angle reward, height reward, obstacle penalty, and velocity reward. 

The mission of the UAV is to approach the target as close as possible, which means that the actions that make the UAV's position closer to the target should be rewarded. Inspired by the references \cite{7,23}, the position reward is designed as:
\begin{equation}
r_{\mathrm{p}}=\frac{D_{\mathrm{pre}}-D_{\mathrm{cur}}}{k_{\mathrm{p}}},
\label{equ20}
\end{equation}
where $D_{\mathrm{pre}}$ and $D_{\mathrm{cur}}$ are the distances between the UAV and the target at the previous timestep and the current timestep, respectively. $k_{\mathrm{p}}$ is a constant to normalize $r_{\mathrm{p}}$.

To get close to the target as quickly as possible, an angle reward is necessary to ensure that the UAV flies in the direction of the target at all times \cite{48}:

\begin{equation}
r_{\mathrm{a}}=\frac{-\left(\varphi_{\text {uT }}+\theta_{\text {uT }}\right)}{k_{\mathrm{a}}},
\label{equ21}
\end{equation}

where $\varphi_{\mathrm{uT}}$ and $\theta_{\mathrm{uT}}$ are the yaw and pitch angles between the UAV and the target, respectively, which are shown in Fig. \ref{fig3} (c) and (d), and $k_{\mathrm{a}}$ is a constant to normalize $r_{\mathrm{a}}$.

In the battlefield environment, the existence of a multipath effect makes the probability of radar finding low-altitude targets low or even zero. The actions that make the height of the UAV lower should be rewarded:

\begin{equation}
r_{\mathrm{h}}=1-\frac{p_{\mathrm{u}z}}{H_{\text{env}}},
\label{equ22}
\end{equation}
where $H_{\text {env }}$ is the maximum height of the environment.

The UAV should complete its tasks while ensuring its own safety, so actions that make the UAV’s state more dangerous should be punished \cite{49}:
\begin{equation}
r_{\mathrm{d}}=d_{\mathrm{ave}}=\sum_{i=1}^{N_{\mathrm{r}}} \frac{d_{i}}{d_{\mathrm{d}}} / N_{\mathrm{r}},
\label{equ23}
\end{equation}
where $d_{\mathrm{d}}$ is the maximum detection distance of the TAR.

The battlefield environment is changing rapidly, and the UAV usually needs to complete a mission within a limited time, such as attacking when the enemy's radar is jammed. Therefore, the faster the UAV's flight speed, the greater the actions should be rewarded:
\begin{equation}
r_{\mathrm{v}}=\frac{\left\|\boldsymbol{v}_{\mathrm{u}}\right\|}{v_{\mathrm{max}}},
\label{equ24}
\end{equation}
where $v_{\mathrm{max}}$ is the maximum flight speed of the UAV.

In addition to the abovementioned influencing rewards, the UAV will also obtain reward $r_{\mathrm{s}}$ when the mission succeeds (reaching the area covered by the target ) or receive penalty $r_{\mathrm{f}}$ when the mission fails (colliding with obstacles or being out of range). In summary, the complete reward function is as follows:

\begin{equation}
r(\boldsymbol{s}, \boldsymbol{a})\!=\!\left\{\begin{array}{cc}
r_{\mathrm{s}} & \text {succeeds} \\
r_{\mathrm{f}} & \text {fails} \\
\lambda_{1} r_{\mathrm{p}}+\lambda_{2} r_{\mathrm{a}}+\lambda_{3} r_{\mathrm{h}}+\lambda_{4} r_{\mathrm{d}}+\lambda_{5} r_{\mathrm{v}} & \text {every step}
\end{array}\right.,
\label{equ25}
\end{equation}
where $\lambda_{1}, \lambda_{2}, \lambda_{3}, \lambda_{4}$ and $\lambda_{5}$ are constants used to tune the contribution rates of different rewards. Among these five rewards, $r_{\mathrm{d}}$ reflects the distance between the UAV and the obstacles, and it directly determines whether the UAV can avoid the obstacles, so it should be assigned a larger coefficient $\lambda_{4}$.  In fact, $r_{\mathrm{h}}$ and $r_{\mathrm{v}}$ are somehow included in $r_{\mathrm{p}}$. To make the UAV's flight trajectory more in line with mission requirements, $r_{\mathrm{h}}$ and $r_{\mathrm{v}}$ are still considered but assigned smaller coefficients $\lambda_{3}$ and $\lambda_{5}$, respectively. 

\section{Asynchronous Curriculum Experience Replay}
In this section, asynchronous experience replay (AER) is first introduced to overcome the limitations of the PER algorithm. Then, curriculum experience replay (CER) is designed to integrate CL with DRL and address a key problem of the proposed AER. Subsequently, the implementation of the integrated ACER algorithm is presented. Finally, the time complexity of the ACER is discussed.

\subsection{Asynchronous Experience Replay}
\subsubsection{Asynchronous TD Error Updating}

The PER algorithm uses the TD errors of different experiences to give them different priorities and then samples and learns according to the priorities, thereby accelerating the convergence speed of the DRL algorithms. In PER, the priorities of the stored experiences are updated through the learning process; that is, an experience will be given a new priority only when it is sampled; otherwise, the priority will remain unchanged forever. This sampled-updated priority updating method caused only the priorities of $N$ experiences to be determined according to the importance of the experience to the current network when sampled.

For an stored experience $\boldsymbol{e}_{i}$ whose priority has not been updated, its sampling probability is:

\begin{equation}
	P\left(\boldsymbol{e}_{i}\right)=\frac{p_{i}^{\alpha}}{\sum\limits_{j=1}^{D} p_{j}^{\alpha}}=\frac{p_{i}^{\alpha}}{\sum\limits_{j=1}^{i-1} p_{j}^{\alpha}+p_{i}^{\alpha}+\sum\limits_{j=i+1}^{D} p_{j}^{\alpha}}.
	\label{equ26}
\end{equation}

If the priority of $\boldsymbol{e}_{i}$ is updated to the real priority $\hat{p}_{i}^{\alpha}$, its sampling probability is: 

\begin{equation}
	\hat{P}\left(\boldsymbol{e}_{i}\right)=\frac{\hat{p}_{i}^{\alpha}}{\sum\limits_{j=1}^{i-1} p_{j}^{\alpha}+\hat{p}_{i}^{\alpha}+\sum\limits_{j=i+1}^{D} p_{j}^{\alpha}}.
	\label{equ27}
\end{equation}

The difference between the real sampling probability $\hat{P}\left(\boldsymbol{e}_{i}\right)$ and the current sampling probability $P\left(\boldsymbol{e}_{i}\right)$ can be calculated: 

\begin{equation}
	\begin{aligned}
		&\hat{P}\left(\boldsymbol{e}_{i}\right)-P\left(\boldsymbol{e}_{i}\right)\\
		&=\!\frac{\left(\hat{p}_{i}^{\alpha}-p_{i}^{\alpha}\right)\left(\sum\limits_{j=1}^{i-1} p_{j}^{\alpha}+\sum\limits_{j=i+1}^{D} p_{j}^{\alpha}\right)}{\left(\sum\limits_{j=1}^{i-1} p_{j}^{\alpha}\!+\!\hat{p}_{i}^{\alpha}\!+\!\sum\limits_{j=i+1}^{D} p_{j}^{\alpha}\!\right)\!\left(\sum\limits_{j=1}^{i-1} p_{j}^{\alpha}\!+\!p_{i}^{\alpha}\!+\!\sum\limits_{j=i+1}^{D} p_{j}^{\alpha}\!\right)}.
	\end{aligned}
	\label{equ28}
\end{equation}

Since $p_{j}>0$ and $1\ge\alpha\ge0$, when $\hat{p}_{i}>p_{i}$, $\hat{P}\left(\boldsymbol{e}_{i}\right)>P\left(\boldsymbol{e}_{i}\right)$; when $\hat{p}_{i}<p_{i}$, $\hat{P}\left(\boldsymbol{e}_{i}\right)<P\left(\boldsymbol{e}_{i}\right)$. This reveals how the priority distribution of the experiences in the experience pool of PER is unreasonable: some of the non-updated experiences should have a larger sampling probability but are ignored due to their past-given low priorities, while some other experiences with high priorities should have a smaller sampling probability and are sampled even if they have less effect on the current network.

\begin{figure}[htb]
	\centering
	\includegraphics[width=3.4in]{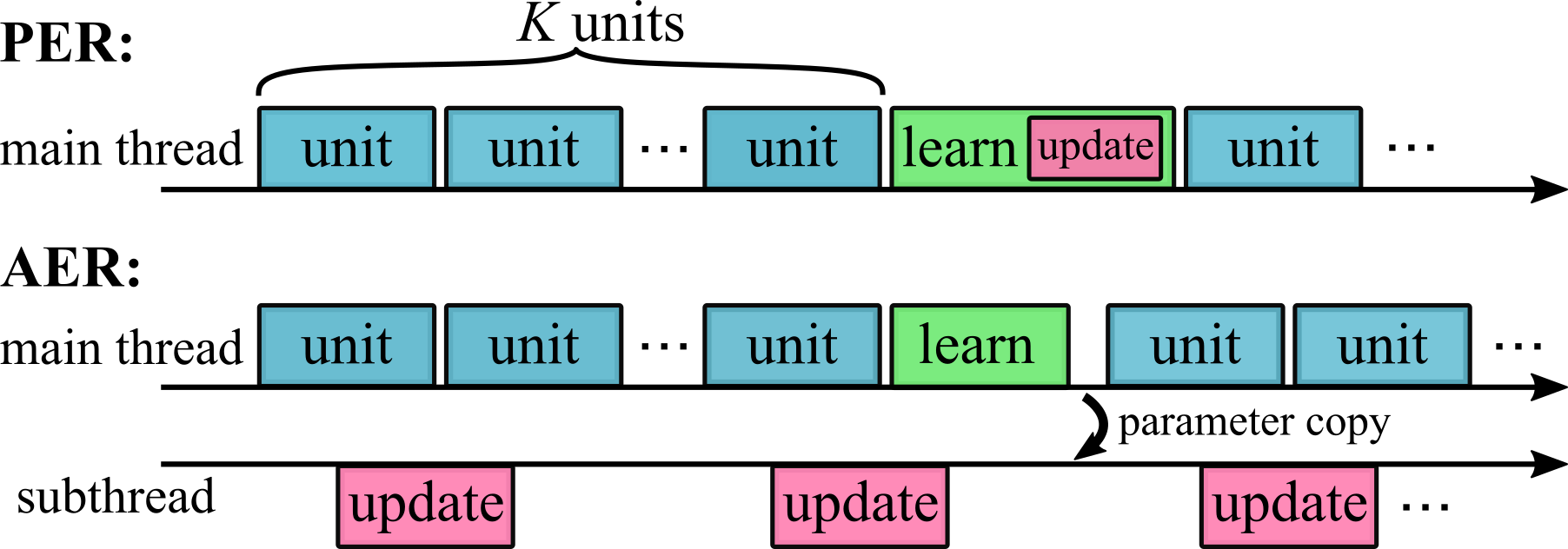}
	\caption{The different training processes of PER and AER. The blue 'unit' module represents the cycle of preforming action and moving to the next state from the current state while receiving a reward. The green 'learn' module contains sampling and updating of the parameters of networks and the pink 'update' module means updating the priorities of specific experiences. The 'parameter copy' means copying the parameters of the main thread's network to the subthread's network. }
	\label{fig4}
\end{figure}

To solve this problem, this paper proposes adopting multithreading technology and opening a subthread to update the priorities of the experiences in the experience pool in order. The main thread and the subthread correspond to a network for avoiding resource access conflicts. As shown in Fig. \ref{fig4}, the proposed AER algorithm separates the updating process from the learning process. The main thread is no longer performing update operations, but only the subthread performs update operations. After each learning, the parameters of the main thread's network are copied to the subthread's network to ensure that the two networks remain consistent. This fast updating method can make the priority distribution of all stored experiences meet the current network requirements as much as possible.
\begin{figure*}[htb]
	\centering
	\includegraphics[width=7.2in]{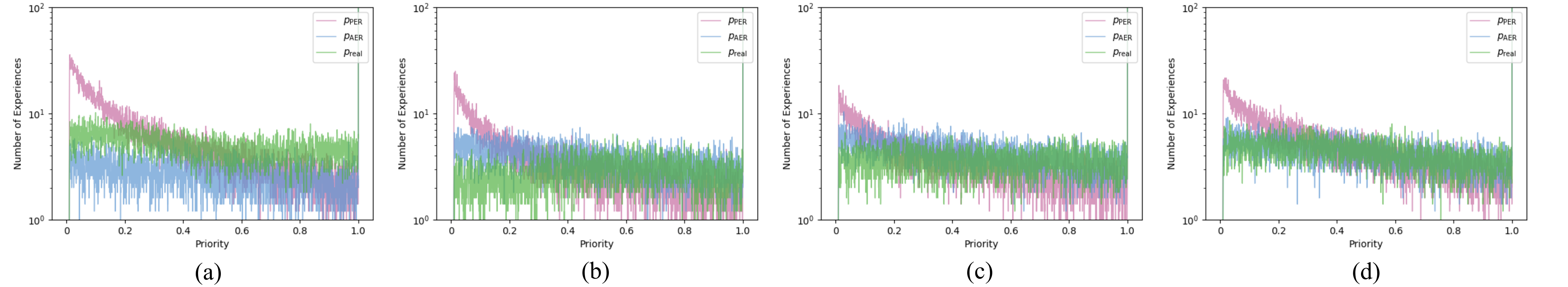}
	\caption{Priority distribution of experiences in experience pools at different training steps. (a). training step = 6e4 (b). training step = 8e4 (c). training step = 10e4 (d). training step = 12e4. $p_\mathrm{PER}$  and $p_\mathrm{AER}$ are the priority distributions of the PER and AER algorithms, respectively, while $p_\mathrm{real}$ is the real priority distribution of the experiences. In this example, the batch size is 16, and the size of the experience pool is 10,000. In addition, the subthred updates the TD errors of 16 stored experiences at every timestep.}
	\label{fig5}
\end{figure*}

A simple experiment is conducted with 'CartPole-v0' \footnote{ http://gym.openai.com/envs/Cartpole-v0.} and the DQN algorithm. The priority distribution of experiences in the entire experience pool at different training stages is generated to show the effectiveness of AER. As shown in Fig. \ref{fig6}, we use $p_\mathrm{real}$, the real priority distribution of all stored experiences at the current timestep, as the standard to measure the accuracy of the priority distributions. At different stages of training, there is a small gap between the priority distributions $p_\mathrm{AER}$ and $p_\mathrm{real}$, and this gap decreases as the networks gradually converge. In comparison, the gap between $p_\mathrm{PER}$ and $p_\mathrm{real}$ is obviously large from beginning to end. This asynchronous TD error updating method of AER allows all experiences to be updated in a short time to obtain the latest priorities. In this way, the gap between the current priority distribution and the $p_\mathrm{real}$ can be narrowed, and the advantages of importance sampling can be given full play to improve the convergence speed of the algorithm.

\subsubsection{True Priority Needs to be Considered}
A good ER algorithm can achieve a balance between the priority and diversity of the sampled experiences \cite{39}. For stability reasons, the PER algorithm clipped the TD errors when updating priorities to fall within $[-1,1]$ \cite{27}. This clip operation ensures the stability of the algorithm and prevents excessive priority from affecting the diversity of sampling. However, it has a great impact on the priority of sampling, that is, all priorities greater than 1 will be clipped to 1, no matter how large it was. It can be seen from Fig. \ref{fig6} that there are many experiences with priority $p=1$ in each period of training, and the probability of these experiences being sampled is completely equal; that is, sampling in this part of the experiences regresses to uniform sampling. There is no doubt that abandoning the true priority will have a huge impact on the quality of the sampled samples, thereby slowing the convergence speed of the algorithm.

The above problem can be solved in two ways: 1) Set different clip ranges for different environments, although doing so will introduce some task-dependent hyperparameters that need careful tuning. 2) Use other methods to replace the clip operation. To maximize the priority while ensuring diversity, this paper abandons the clip operation, assigns experience to the true priority based on its TD error, and proposes a method combined with CL to avoid the appearance of outliers. This method will be introduced in detail in later sections.

\subsubsection{Newest is Useful}

Novel experiences, the experiences that have not been learned, are quite important, because the actual process of DRL is to optimize the policy through learning the stored experiences, thereby generating novel experiences for further learning, and to continuously loop this process until the optimal policy is found. Novel experiences can be divided into two categories: 1) a novel state with any action and 2) an old state with a novel action. For the first case, there is no doubt that these novel experiences are more worth learning. The agent has never reached these states, so it has not learned what motion policy should be taken at these states. Learning from these novel experiences can improve the agent's policy and promote the agent to explore a wider range of the environment. The generation of this kind of novel experience mainly comes after the second kind of novel experience: the agent executes novel actions at the old state to reach novel states.

Assuming that at any timestep during the training process, for a state $\boldsymbol{s}$, $\boldsymbol{a}_{1}$ is the optimal action. The agent should perform action $\boldsymbol{a}_{1}$ at state $\boldsymbol{s}$  and generate the old experience $\left(\boldsymbol{s}, \boldsymbol{a}_{1}, r_{1}, \boldsymbol{s}_{1}\right)$. However, due to the existence of $\varepsilon$-greedy and exploration noise, the agent chooses a novel action $\boldsymbol{a}_{2}$  and gains novel experience $\left(\boldsymbol{s}, \boldsymbol{a}_{2}, r_{2}, \boldsymbol{s}_{2}\right)$. Taking the DQN algorithm as an example, the TD errors of these two experiences are:
\begin{equation}
	\begin{aligned}
		&\delta_{1}=r_{1}+\gamma \max _{\boldsymbol{a}_{1}^{\prime}} Q^{\prime}\left(\boldsymbol{s}_{1}, \boldsymbol{a}_{1}^{\prime}\right)-Q\left(\boldsymbol{s}, \boldsymbol{a}_{1}\right), \\
		&\delta_{2}=r_{2}+\gamma \max _{\boldsymbol{a}_{2}^{\prime}} Q^{\prime}\left(\boldsymbol{s}_{2}, \boldsymbol{a}_{2}^{\prime}\right)-Q\left(\boldsymbol{s}, \boldsymbol{a}_{2}\right).
	\end{aligned}
	\label{equ29}
\end{equation}

For the problem of high-dimensional and large-scale state space such as UAV AMC, the change of state $\boldsymbol{s}$ by an action $\boldsymbol{a}$ in a timestep can be ignored, so we have $\max _{\boldsymbol{a}_{i}^{\prime}} Q^{\prime}\left(\boldsymbol{s}^{\prime}, \boldsymbol{a}_{i}^{\prime}\right) \approx \max _{\boldsymbol{a}} Q^{\prime}(\boldsymbol{s}, \boldsymbol{a})$. Assuming $\delta_{i}>0$, we can get
\begin{equation}
	\delta_{2}-\delta_{1} \approx r_{2}-r_{1}+Q\left(\boldsymbol{s}, \boldsymbol{a}_{1}\right)-Q\left(\boldsymbol{s}, \boldsymbol{a}_{2}\right).
	\label{equ30}
\end{equation}

1) When $\boldsymbol{a}_{2}$ is a better action, that is, $r_{1}<r_{2}$:

Since $\boldsymbol{a}_{1}$ is the optimal action until now and the novel experience $\left(\boldsymbol{s}, \boldsymbol{a}_{2}, r_{2}, \boldsymbol{s}_{2}\right)$ has not been learned, we have
\begin{equation}
	Q\left(\boldsymbol{s},\boldsymbol{a}_{1}\right)=\max _{\boldsymbol{a}} Q(\boldsymbol{s}, \boldsymbol{a})>Q\left(\boldsymbol{s}, \boldsymbol{a}_{2}\right),
	\label{equ31}
\end{equation}
and $\delta_{2}>\delta_{1}$.

2) When $\boldsymbol{a}_{2}$ is a worse action, that is, $r_{1}>r_{2}$: 

The DQN algorithm uses a neural network to update the Q value, where each update of $Q(\boldsymbol{s}, \boldsymbol{a})$ can still be expressed by
\begin{equation}
	Q(\boldsymbol{s}, \boldsymbol{a})\!=\!Q(\boldsymbol{s}, \boldsymbol{a})\!+\!\alpha\!\left[r\!+\!\gamma \max _{\boldsymbol{a}^{\prime}} Q^{\prime}\!\left(\boldsymbol{s}^{\prime}, \boldsymbol{a}^{\prime}\right)\!-\!Q(\boldsymbol{s}, \boldsymbol{a})\right].
	\label{equ32}
\end{equation}

However, due to the parameter update of the neural network, the update of a $Q$ value will have a certain impact on other $Q$ values. For a novel experience $\left(\boldsymbol{s}, \boldsymbol{a}_{2}, r_{2}, \boldsymbol{s}_{2}\right)$ that has not been learned before, its $Q$ value is also constantly changing in the past learning process.

Here, we make a reasonable assumption that before the training starts, the probability of all actions being selected in any state $\boldsymbol{s}$ is equal, that is, $Q_{0}\left(\boldsymbol{s}, \boldsymbol{a}_{1}\right)=Q_{0}\left(\boldsymbol{s}, \boldsymbol{a}_{2}\right)$. Suppose that by the time the novel experience appears, the old experience $\left(\boldsymbol{s}, \boldsymbol{a}_{1}, r_{1}, \boldsymbol{s}_{1}\right)$ has been learned $n$ times; then, $Q\!\left(\boldsymbol{s}, \boldsymbol{a}_{1}\right)\!-\!Q\!\left(\boldsymbol{s}, \boldsymbol{a}_{2}\right)$ can be estimated by (\ref{equ33}).
\newcounter{mytempeqncnt}
\begin{figure*}[!t]
\setcounter{equation}{32}	
\begin{equation}
	\begin{aligned}
		&\ \ \ Q\!\left(\boldsymbol{s}, \boldsymbol{a}_{1}\right)\!-\!Q\!\left(\boldsymbol{s}, \boldsymbol{a}_{2}\right) \\
		&\!=\!Q_{n}\!\left(\boldsymbol{s}, \boldsymbol{a}_{1}\right)\!-\!Q_{n}\!\left(\boldsymbol{s}, \boldsymbol{a}_{2}\right) \\
		&\!\approx\!\left\{\!Q_{n-1}\!\left(\boldsymbol{s}, \boldsymbol{a}_{1}\right)\!+\!\alpha\!\left[\!r_{1}+\gamma \max _{\boldsymbol{a}_{1}^{\prime}} Q_{n-1}^{\prime}\!\left(\boldsymbol{s}_{1}, \boldsymbol{a}_{1}^{\prime}\right)\!-\!Q_{n-1}\!\left(\boldsymbol{s}, \boldsymbol{a}_{1}\right)\!\right]\!\right\}\!-\!\left\{\!Q_{n-1}\left(\!\boldsymbol{s}, \boldsymbol{a}_{2}\right)\!+\!\alpha\!\left[\!r_{2}+\gamma \max _{\boldsymbol{a}_{2}^{\prime}} Q_{n-1}^{\prime}\left(\!\boldsymbol{s}_{2}, \boldsymbol{a}_{2}^{\prime}\right)\!-\!Q_{n-1}\!\left(\boldsymbol{s}, \boldsymbol{a}_{2}\right)\!\right]\!\right\}\! \\
		&\!=\!(1-\alpha)\left[Q_{n-1}\left(\boldsymbol{s}, \boldsymbol{a}_{1}\right)-Q_{n-1}\left(\boldsymbol{s}, \boldsymbol{a}_{2}\right)\right]+\alpha\left(r_{1}-r_{2}\right) \\
		&\!=\!(1-\alpha)\left\{(1-\alpha)\left[Q_{n-2}\left(\boldsymbol{s}, \boldsymbol{a}_{1}\right)-Q_{n-2}\left(\boldsymbol{s}, \boldsymbol{a}_{2}\right)\right]+\alpha\left(r_{1}-r_{2}\right)\right\}+\alpha\left(r_{1}-r_{2}\right) \\
		&\!=\!(1-\alpha)^{2}\left[Q_{n-2}\left(\boldsymbol{s}, \boldsymbol{a}_{1}\right)-Q_{n-2}\left(\boldsymbol{s}, \boldsymbol{a}_{2}\right)\right]+[\alpha+\alpha(1-\alpha)]\left(r_{1}-r_{2}\right) \\
		&\!=\!(1-\alpha)^{n}\left[Q_{0}\left(\boldsymbol{s}, \boldsymbol{a}_{1}\right)-Q_{0}\left(\boldsymbol{s}, \boldsymbol{a}_{2}\right)\right]+\left[\alpha+\alpha(1-\alpha)+\cdots+\alpha(1-\alpha)^{n-1}\right]\left(r_{1}-r_{2}\right) \\
		&\!=\!\left[1-(1-\alpha)^{n}\right]\left(r_{1}-r_{2}\right)
	\end{aligned}
	\label{equ33}
\end{equation}
\hrulefill
\vspace*{4pt}
\end{figure*}

Thus, the difference in TD errors of the novel and the old experience is  
\begin{equation}
	\begin{aligned}
		\delta_{2}-\delta_{1} 
		&\approx r_{2}-r_{1}+Q\left(\boldsymbol{s}, \boldsymbol{a}_{1}\right)-Q\left(\boldsymbol{s}, \boldsymbol{a}_{2}\right) \\
		&=-(1-\alpha)^{n}\left(r_{1}-r_{2}\right).
	\end{aligned}
	\label{equ34}
\end{equation}	 

Since $0<\alpha<1$, we can get
\begin{equation}
	-(1-\alpha)^{n}<0,
	\label{equ35}
\end{equation}
and $\delta_{2}<\delta_{1}$.

In summary, when the novel action is a better action, the novel experience is more worth learning than the old experience.

New experiences are generated under the guidance of the latest policy, and they are more likely to contain novel experiences. Combined experience replay \cite{36} replays the newest experience every learning time to improve the performance of DQN with a large experience pool. PER \cite{27} assigns the newest experiences with the greatest priority among all stored experiences. However, combined experience replay only replays the newest experience and PER cannot guarantee that the newest experiences can be learned at least once; neither of these two advanced ER algorithms can take full advantage of novel experiences. 

\begin{figure}[H]
	\centering
	\includegraphics[width=3.5in]{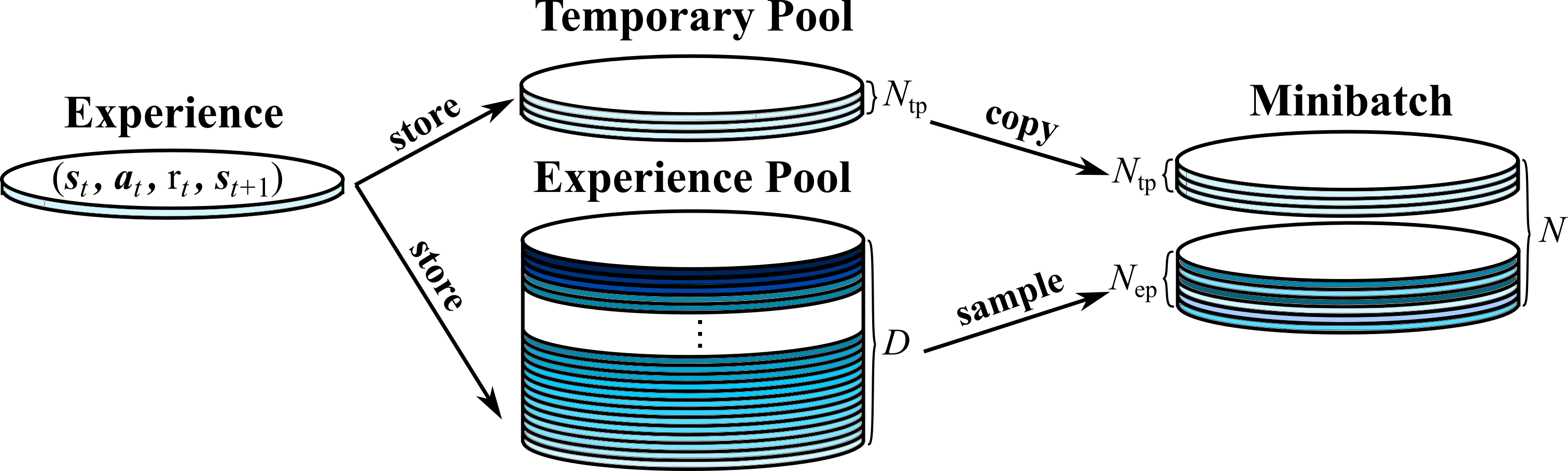}
	\caption{Experience storing and sampling of AER. An experience is represented by a slice with different colors. The difference in color indicates that the experience is different from the old to the new. The new experience is lighter in color and the old experience is darker in color.}
	\label{fig6}
\end{figure}

To remedy this situation, this paper constructed a small FIFO experience pool named the temporary pool to temporarily store the serial number of the newest experiences (Fig. \ref{fig6}). Every new experience will be stored in the experience pool and its serial number will be stored in the temporary pool. When sampling, the $N_\mathrm{tp}$ experiences copied according to the serial numbers stored in the temporary pool and the $N_\mathrm{ep}$ experiences sampled from the remaining experiences in the experience pool together form a minibatch of size $N$ for updating networks. This temporary pool ensures that the same experience will not be sampled multiple times during one sampling, thereby effectively avoiding overfitting and making full use of the newest experiences. A deep survey of the influence of different $N_\mathrm{tp}$ on the algorithm will be presented in Section V.

\subsubsection{Oldest is Not the Worst}
To the best of our knowledge, all the experience pools of the off-policy DRL algorithms are FIFO buffers and there is no research trying to use other buffer forms to store experiences. This paper explores and designs a first-in-useless-out (FIUO) buffer to improve the quality of the experience stored in the experience pool to achieve better algorithm performance. The adoption of FIFO buffer is because it is generally believed that the older the experience, the less learning value it contains. This may not be the case, because old experiences are not necessarily the worst.

This paper believes that even if some of the older experiences do not have much effect on updating the current networks, they may still play a key role at some point in the future. Compared with some older experiences, some new experiences with less value for the current network update should be replaced. In AER, when a new experience needs to be stored, if the experience pool is full, the experience with the lowest priority will be replaced.

We improved the 'sum-tree' of the PER algorithm to form a FIUO buffer named 'double sum-tree' to reduce the computational complexity of finding the most useless experience in the entire experience pool. Due to the asynchronous TD error updating subthread, the priorities of all stored experiences are constantly being updated, so it can be ensured that the replaced experience is more useless in a short time.
\begin{figure}[htb]
	\centering
	\includegraphics[width=1.8in]{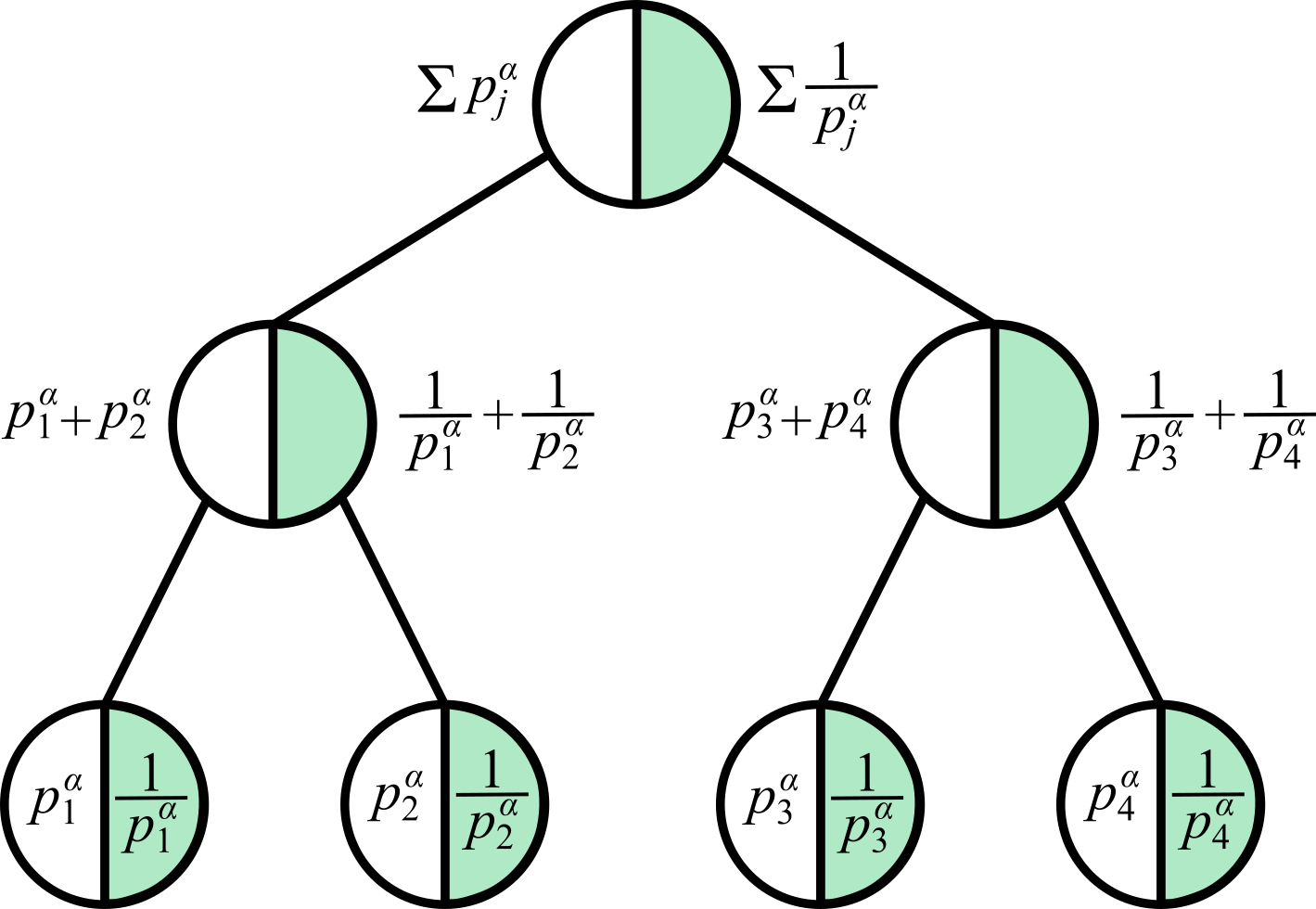}
	\caption{The 'double sum-tree' data structure. Each node stores two data sampling factors (shown in the white semicircle) and replacing factors (shown in the green semicircle). The sampling factors are used for sampling experiences to update policy, while the replacing factors are used for finding more useless experiences to be replaced when new experiences need to be stored.}
	\label{fig7}
\end{figure}

As shown in Fig. \ref{fig7}, unlike 'sum-tree', which only stores the sampling factor $p_{i}^\alpha$ of experiences $\boldsymbol{e}_{i}$ in the leaf node, 'double sum-tree' also stores the replacing factor $\frac{1}{p_{i}^{\alpha}}$. The parent node also has two data points $\sum\limits{p_{j}^{\alpha}}$ and $\sum\limits\frac{1}{p_{j}^{\alpha}}$, which are the sum of the sampling factors and replacing factors of its child nodes, respectively. The probability of replacing experience $\boldsymbol{e}_{i}$ is

\begin{equation}
	P_{\mathrm{rep}}\left(\boldsymbol{e}_{i}\right)=\frac{\frac{1}{p_{i}^{\alpha}}}{\sum\limits_{j=1}^{D}\frac{1}{p_{j}^{\alpha}}}.
	\label{equ36}
\end{equation}
As a result, by using this 'double sum-tree', the AER can make full use of the older experiences while ensuring the efficiency of the algorithm.

\subsection{Curriculum Experience Replay}
Learning from simple to difficult is the universal order of human learning knowledge because simple knowledge lays the foundation for more difficult future learning. Motivated by this human learning process, Bengio \cite{29} proposed CL for machine learning, which advocates letting the model learn from easier samples and gradually increasing the difficulty of the samples. Due to the two benefits of less training time and better model generalization ability brought by CL, CL has been widely used in computer vision \cite{50}, natural language processing \cite{51}, DRL \cite{52,53} and other fields. All these applications demonstrate that as a flexible plug-and-play submodule independent of the original training, CL is easy to use to speed up the learning process, especially for deep neural networks \cite{54}. This paper is the first to attempt to introduce CL to prescribe the order of the experiences learned by the UAV to better solve the AMC problem.

\begin{figure}[htb]
	\centering
	\includegraphics[width=3.4in]{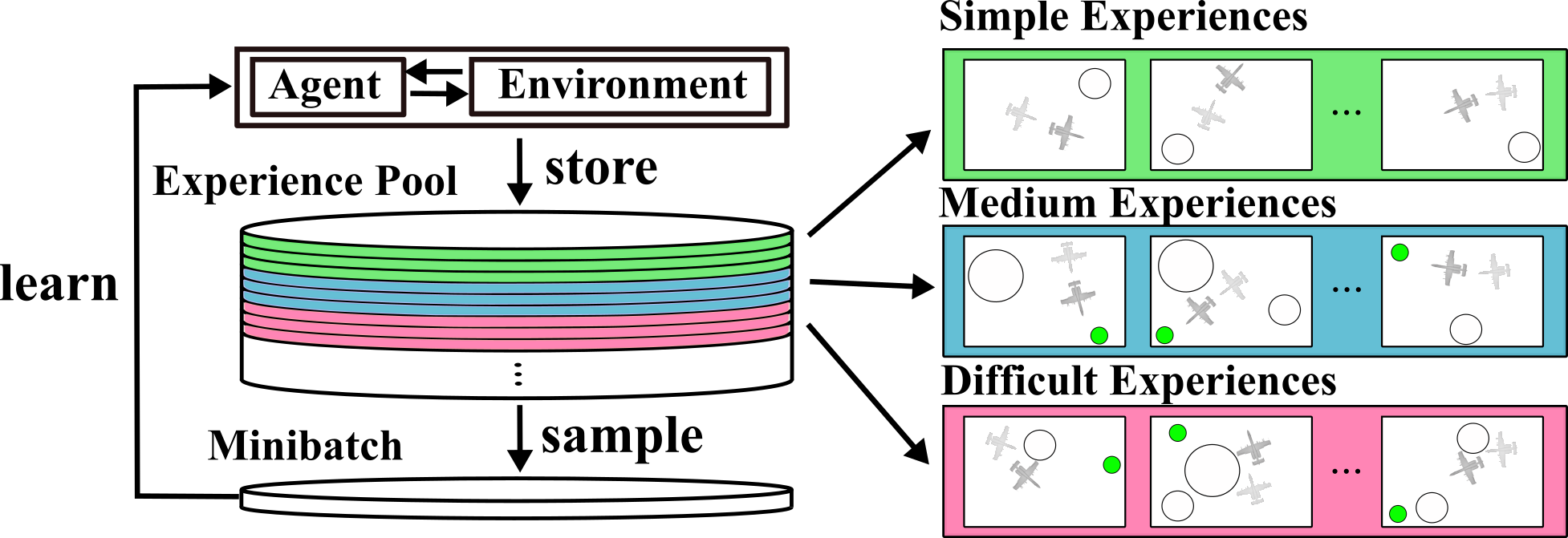}
	\caption{Framework of CER. Each experience $\left(\boldsymbol{s}_{t}, \boldsymbol{a}_{t}, r_{t}, \boldsymbol{s}_{t+1}\right)$ is presented by a 2D picture. The white circles and the green circle represent obstacles and target, respectively. The state $\boldsymbol{s}_{t}$ is represented by the translucent UAV, and the state $\boldsymbol{s}_{t+1}$ is represented by the opaque UAV. Different colors represent different levels of experience difficulties.}
	\label{fig8}
\end{figure}

The first task of combining CL and ER is to define a reasonable standard to measure the difficulty of different experiences. For example, the experiences of UAVs can be simply divided into three categories according to difficulty (Fig. \ref{fig8}): 1) steady forward flight (simple experiences); 2) steady flight toward the target (medium experiences); and 3) steady flight toward the target and avoid obstacles (difficult experiences). However, UAV AMC in dynamic unknown environments is a complex problem, which means that artificially assigning the difficulties of experiences may be inappropriate and that more professional classification standards should be applied. Motivated by the PER algorithm, which assigns different experiences with different priorities $p$ according to the TD errors, we find that the TD error $\delta$ is a good standard for dividing the difficulties of experiences. In deep neural networks, transitions with large magnitudes of TD errors require a smaller step size to follow the curvature of the objective function \cite{55}. The larger the $\delta$ is, the greater the impact of the experience on the current network, the more the experience should be learned, and, in a sense, the greater the difficulty of fully grasping the policy of the experience.

\begin{figure}[h]
	\centering
	\includegraphics[width=3.5in]{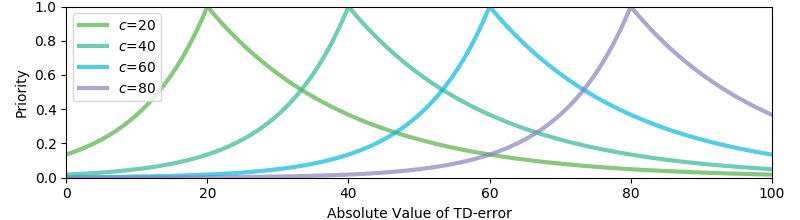}
	\caption{Priority function with different $c$ ($k_1=0.1$ and $k_2=0.05$).}
	\label{fig9}
\end{figure}

A priority function for measuring the difficulty of different experiences should meets the following conditions: 1) The priority function should be a bounded function in $[0,+\infty)$ to ensure that the priority difference between experiences will not be too large to avoid the occurrence of outliers. 2) The value range of the priority function should be greater than 0 to ensure that every experience has the probability of being sampled. 3) The priority function should increase monotonically first and then decrease monotonically to ensure that the experience corresponding to the curriculum factor can have the greatest priority. 4) The slope of the priority function should be easily adjusted to ensure that the priority function can be adjusted to suit with different environments. The first two conditions can ensure the diversity of the sampled experiences, while the third and fourth conditions can ensure the sampling priority and the universality of the priority function, respectively. After a certain amount of exploration, the priority function of CER is depicted as follows:
   
\begin{equation}
p(\delta, c)= \begin{cases}\exp \left(k_{1} \cdot(|\delta|-c)\right) & |\delta| \leq c \\ \exp \left(k_{2} \cdot(c-|\delta|)\right) & |\delta|>c\end{cases},
\label{equ37}
\end{equation}
where $c$ is the curriculum factor that indicates the learning stages, and $k_1$ and $k_2$ are constants used to adjust the slope of the priority function. Fig. \ref{fig9} gives a sketch of the priority function with different $c$.

By using this well-designed priority function whose value range is $(0,1]$, the clip problem introduced by the true TD error can be solved. In the whole training process, the initial value of $c$ is $c_\mathrm{init}$, which is updated at regular intervals by $c=c+c_{\mathrm{incr}}$. Unlike the DCRL algorithm updates the curriculum factor at each learning step, we believe that an update frequency that is too high causes the agent to be unable to learn steadily because the difficulty of the experiences to be learned is constantly changing. The CER algorithm updates $c$ every $U_\mathrm{c}$ episodes to ensure that the agent has a consistent learning standard in every $U_\mathrm{c}$ episodes. In addition, after exploration, we found that $k_1>k_2$ should be set to increase the priorities of experiences with greater difficulties, which can have a positive impact on the convergence of the network.

\begin{figure}[htb]
	\centering
	\includegraphics[width=3.5in]{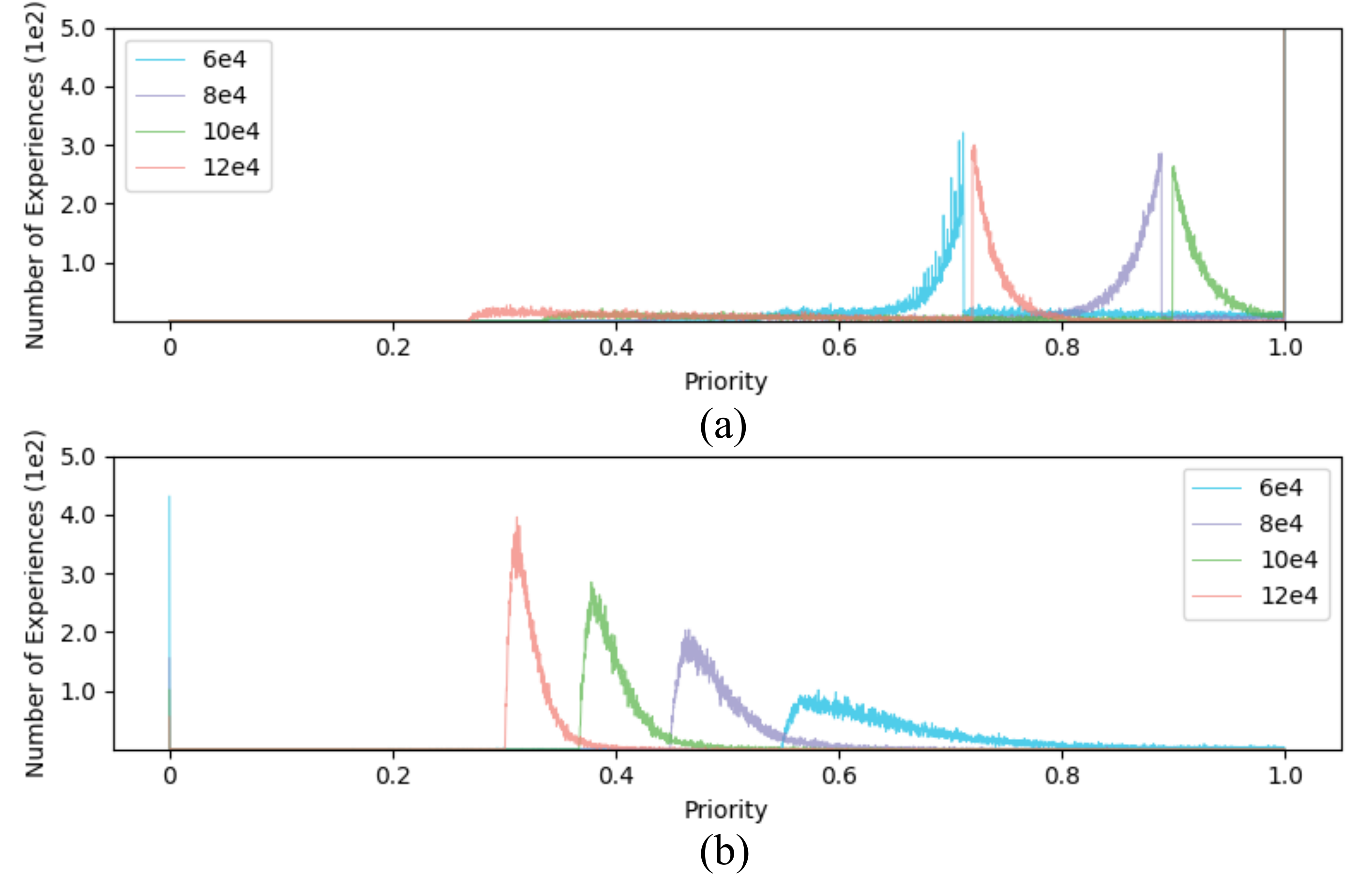}
	\caption{Priority distribution of experiences in experience pools at different training steps. (a). CER. (b). AER+CER.
		To compare with the previous experiment, the same hyperparameters are used, and the learning steps are also selected as 6e4, 8e4, 10e4, and 12e4.}
	\label{fig10}
\end{figure}

To verify the impact of CER, experiments are conducted on 'CartPole-v0', and the results are shown in Fig. \ref{fig10}. It can be clearly seen that the CER algorithm effectively changes the priority distribution of experiences in the experience pool. As we still assign the maximum priority of all experiences to the new experiences, there are still many experiences with priority 1 when the CER is applied alone (Fig. \ref{fig10}(a)). When the AER is added, the subthread quickly updates the priorities of all experiences in the experience pool, and these experiences with priority 1 are given their true priorities. In addition, following the curriculum factor $c$, the priority distribution of experience in the experience pool becomes the distribution we want, as shown in Fig. \ref{fig10}(b). Under this distribution, the agent can better choose the experience that suits its current state to learn and achieve rapid and stable convergence of the policy.

\subsection{Implementation}
AER can accelerate the formation of a curriculum priority distribution of experiences and the conversion between curriculum priority distributions of different difficulties to ensure that the agent can learn stably according to the difficulty of the experiences. CER can enable AER to give true priority to experiences to make full use of the advantages of importance sampling, thereby accelerating the speed of agent learning. AER and CER complement each other, and finally form the ACER algorithm. We applied the ACER to the TD3 algorithm to show its performance and present the pseudocode of the ACER-TD3 algorithm (Algorithm \ref{alg1}).
\begin{algorithm}[htb] 
	\scriptsize
	\caption{ACER-TD3} 
	\begin{algorithmic}
		\REQUIRE experience pool $\mathrm{R}$ with the capacity $D$, temporary pool $\mathrm{R_{tp}}$ with the capacity $N_\mathrm{tp}$, minibatch $N$, replay period  $K$, actor update delay $d$, training episode $M$, the number of asynchronous TD error update experiences $A$, update period $U_\mathrm{c}$, curriculum factor $c$ and exponents $\alpha$ and $\beta$.
		\REQUIRE critics $Q_{1}\left(\boldsymbol{s}, \boldsymbol{a} \mid \boldsymbol{\theta}^{Q_{1}}\right)$, $ Q_{2}\left(\boldsymbol{s}, \boldsymbol{a} \mid \boldsymbol{\theta}^{Q_{2}}\right)$, and actor $\mu\left(\boldsymbol{s} \mid \boldsymbol{\theta}^{\mu}\right)$ with weights $\boldsymbol{\theta}^{Q_{1}}$, $\boldsymbol{\theta}^{Q_{2}}$, and $\boldsymbol{\theta}^{\mu}$.
		\REQUIRE target nets $Q_{1}^{\prime}, Q_{2}^{\prime}$, and $\mu^{\prime}$ with weights $\boldsymbol{\theta}^{\prime} \leftarrow \boldsymbol{\theta}$.
		\FOR{$episodes=1$, $M$}
		\FOR{$t=1$, $T$}
		\STATE Select action with exploration noise $\boldsymbol{a}_{t}=\mu\left(\boldsymbol{s}_{t} \mid \boldsymbol{\theta}^{\mu}\right)+\varepsilon, \varepsilon \sim \mathcal{N}(0, \sigma)$ and observe the reward $r_{t}$ and new state $\boldsymbol{s}_{t+1}$.
		\STATE Store the experience $\left(\boldsymbol{s}_{t}, \boldsymbol{a}_{t}, r_{t}, \boldsymbol{s}_{t+1}\right)$ in the temporary pool $\mathrm{R_{tp}}$ and the experience pool $\mathrm{R}$ with maximal priority $p_t=\max p_{i}$.
		\IF{$t \equiv 0 \bmod K$}
		\STATE Sample $N_\mathrm{ep}$ experiences from $\mathrm{R}$ according to the sample probability $i \!\sim\! P(i)\!=\!p_{i}^{\alpha} / \sum_{j} p_{j}^{\alpha}$ combined with the $N_\mathrm{tp}$ temporary experiences of $\mathrm{R_{tp}}$ to form the minibatch $\left(\boldsymbol{s}_{i}, \boldsymbol{a}_{i}, r_{i}, \boldsymbol{s}_{i+1}\right)_{i=1 \ldots N}$.
		\STATE Compute the importance-sampling weight:
		\STATE 
		\begin{center}
			$\omega_{i}=(D \cdot P(i))^{-\beta} / \max _{j} \omega_{j}$.
		\end{center}
		\STATE Set:
		\STATE $y_{i}=r\left(\!\boldsymbol{s}_{i}, \boldsymbol{a}_{i}\!\right)+\gamma \min _{j=1,2} Q_{j}^{\prime}\left(\!\boldsymbol{s}_{i+1}, \mu^{\prime}\left(\!\boldsymbol{s}_{i+1} \!\mid\! \boldsymbol{\theta}^{\mu^{\prime}}\!\right)+\varepsilon \!\mid\! \boldsymbol{\theta}^{Q_{j}^{\prime}}\!\right)$, 
		\STATE 
		\begin{center}
			$\varepsilon \sim \operatorname{clip}(\mathcal{N}(0, \tilde{\sigma}),-l, l)$.
		\end{center}
		\STATE Update critic by minimizing the loss:
		\STATE 
		\begin{center}
			 $L\left(\boldsymbol{\theta}^{Q_{j}}\right)=\min _{j=1,2} \frac{1}{N} \sum_{i}\omega_{i}\cdot\left(y_{i}-Q_{j}\left(\boldsymbol{s}_{i}, \boldsymbol{a}_{i} \mid \boldsymbol{\theta}^{Q_{j}}\right)\right)^{2}$.
		\end{center}
		\IF{$t \equiv 0 \bmod d$}
		\STATE Update the actor policy using the sampled policy gradient:
		\STATE			$\nabla_{\boldsymbol{\theta}^{\mu}}\! J\! \approx \!\frac{1}{N} \!\sum_{i} \!\nabla_{\boldsymbol{a}} Q\!\left(\!\boldsymbol{s}, \boldsymbol{a}\! \mid\! \boldsymbol{\theta}^{Q}\!\right) \!\mid\! _{\boldsymbol{s}=\boldsymbol{s}_{i}, \boldsymbol{a}=\mu\!\left(\boldsymbol{s}_{i}\right)}\! \nabla_{\boldsymbol{\theta}^{\mu}} \mu\left(\!\boldsymbol{s}\!\mid\! \boldsymbol{\theta}^{\mu}\!\right)\!\mid\! {\boldsymbol{s}_{i}}$.
		\ENDIF
		\STATE Update the target networks: $\boldsymbol{\theta}^{\prime} \leftarrow \tau \boldsymbol{\theta}+(1-\tau) \boldsymbol{\theta}^{\prime}$.
		\STATE Copy network parameters from main thread to subthread.
		\ENDIF
		\STATE Subthread samples $A$ experiences from $\mathrm{R}$.
		\STATE Compute the TD error:
		\STATE
		\begin{center}
			$\delta_{i}=y_{i}-Q_{j}\left(\boldsymbol{s}_{i}, \boldsymbol{a}_{i} \mid \boldsymbol{\theta}^{Q_{j}}\right)$.
		\end{center}
		\STATE Update the experience priority:
		\STATE
		\begin{center}
			$p_{i} \leftarrow p\left(\delta_{i}, c\right)= \begin{cases}\exp \left(k_{1} \cdot\left(\left|\delta_{i}\right|-c\right)\right) & \left|\delta_{i}\right| \leq c \\ \exp \left(k_{2} \cdot\left(c-\left|\delta_{i}\right|\right)\right) & \left|\delta_{i}\right|>c\end{cases}$.
		\end{center}
		\ENDFOR
		\IF{$episodes \bmod U_\mathrm{c}$}
		\STATE Update curriculum factor $c=c+c_{\text {incr }}$.
		\ENDIF
		\ENDFOR		
	\end{algorithmic}
\label{alg1}
\end{algorithm}

\subsection{Analysis of the Algorithm Complexity}
The ACER algorithm uses a subthread to asynchronously update the TD errors of $A$ experiences costs $O(A)$ each timestep. The operation of assigning true priority of the new experience and storing it in the temporary pool both increase the time complexity of $O(1)$. In addition, finding the experience to be replaced in the FIUO buffer takes $O(\mathrm{log}D)$. In summary, the time complexity of ACER increased by $O(A)$ compared to the PER algorithm at each timestep. Since the asynchronous update operation is completed by the subthread, the running time of the ACER algorithm will not increase significantly.

\section{Experiments}

\subsection{Settings}

\begin{figure}[htb]
	\centering
	\includegraphics[width=3.5in]{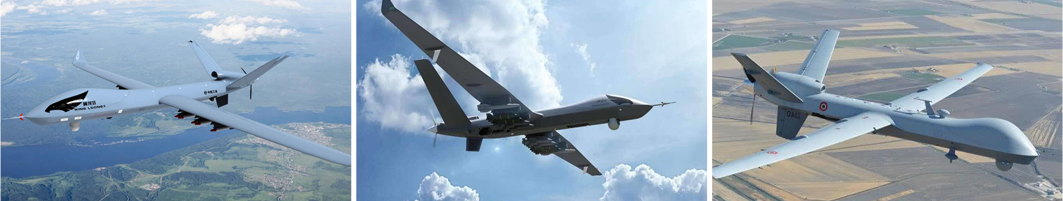}
	\caption{Chinese ‘Wing Loong II’ UAV.
	}
	\label{fig11}
\end{figure}

To simulate the realistic UAV state as much as possible, a UAV model based on the parameters of the ‘Wing Loong II’ UAV (Fig. \ref{fig11}), an identify and destroy integrated UAV developed by the Chinese Chengdu Aircraft Design and Research Institute, is constructed. The maximum flight speed and the maximum load factor of the UAV are set to 103$\mathrm{~m/s}$ and 15, respectively. For the TAR of the UAV, $N_\mathrm{r}$ is 32 to return the environmental state, and the detection distance is set to 5$\mathrm{~km}$.

\begin{figure}[htb]
	\centering
	\includegraphics[width=3.5in]{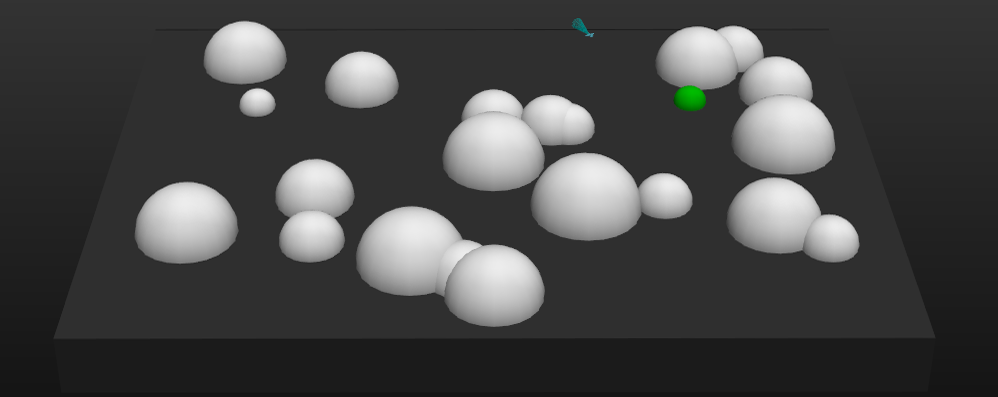}
	\caption{The Simulation Environment.}
	\label{fig12}
\end{figure}

For the simulation of the battlefield, a large-scale dynamic 3D environment ranging from $120 \times 90 \times 10 \mathrm{~km}^{3}$ is designed to ensure UAVs perform different missions. As shown in Fig. \ref{fig12}, the simulation environment mainly contains three modules: the UAV, the target, and the obstacles. The target, represented by the green hemisphere, is set with a fixed radius of 3$\mathrm{~km}$, while the obstacles are represented by the white hemispheres whose radii range from 5-10$\mathrm{~km}$. The initial positions of the target and UAV are randomly generated, and the distance between them is greater than 50$\mathrm{~km}$ to increase the exploration of every episode. The direction of movement of each obstacle is also random. To more easily calculate the state information of the UAV, the acceleration of gravity is a fixed value of 9.8$\mathrm{~m/s}^{2}$. In addition, after much exploration and experimentation, the reward function is set as follows: the rewards $r_s$ and  $r_f$ are 100 and -200, respectively; the contribution rate factors of different rewards are $\lambda_{1}\!=\!20$, $\lambda_{2}\!=\!20$, $\lambda_{3}\!=\!10$, $\lambda_{4}\!=\!40$, and $\lambda_{5}\!=\!10$. More information about the simulation environment can be found in our previous work \cite{12}. 

In addition to the proposed ACER algorithm, two baseline algorithms of TD3 \cite{30} and PER \cite{27} and two novel ER algorithms REL \cite{12} and DCRL \cite{28} are trained for comparison. For the hyperparameters of the TD3 algorithm, both the actor network and the two critic networks contain two hidden layers with 100 nodes each. The Adam optimizer is employed to update the network parameters with learning rates of 0.0001 and 0.001 for the actor and critic, respectively. The discount factor $\gamma$ is 0.9. The soft update rates $\tau$ of the actor and critic are set to 0.1 and 0.2, respectively. Both the action exploration noise and the target policy smoothing noise satisfy a Gaussian distribution  $\mathcal{N}(0,0.1)$. The actor update delay $d$ is 2. For the ER of the training process, the capacity $D$ of the experience pool is 50,000. The size of the minibatch $N$ is 256. The replay period $K$ is 20. The other hyperparameters of the ER part of the PER, REL, and DCRL algorithms are the same as those in \cite{12}, \cite{27}, and \cite{28}. For the proposed ACER algorithms, the exponent $\alpha$ is 0.6 and the importance sample variable $\beta$ is linearly annealed from 0.4 to 1. The number of asynchronous TD error updating experiences $A$ is 256. The capacity of the temporary experience pool $N_\mathrm{tp}$ is 5. The initial value of curriculum factor $c$ is 10, and the increment of curriculum factor $c_\mathrm{incr}$ is 1. The curriculum factor update period $U_\mathrm{c}$ is 100. The $k_1$ and $k_2$ of the priority function are 0.01 and 0.005, respectively. In addition, every agent is warmed up by an initial 200 episodes without training. The total number of episodes for each training is 5,000, and the maximum timestep of each episode is 3,000. Each training takes about 10 hours. All the experiments are carried out on a computer with an Nvidia RTX 2080Ti GPU, Ubuntu 16.04 LTS, and Python.

\subsection{Experimental Results}

To validate the efficiency of the proposed ACER algorithm, an environment containing 20 random obstacles with a speed of 5$\mathrm{~m/s}$ is set to train the above five algorithms. The hit rate, the probability of the UAV successfully hitting the target in the last 500 episodes, is defined as a main evaluation indicator. In addition, the following indicators are defined from different perspectives to further compare the pros and cons of the algorithms: 1) Training Peak (TP):  The peak hit rate of the trained agent during the entire 5,000 episodes of training. 2) Convergence Time (CT): The number of episodes where the hit rate first reached 70\%. 3) Stability after Convergence (SC): The standard deviation of the hit rate in the last 1,500 episodes. 4) Convergence Result (CR): The average hit rate in the last 1,500 episodes.
Each algorithm is trained 10 times under different random seeds to obtain the average value to avoid the influence of random numbers. The complete experimental results are shown in Fig.\ref{fig13} and Table \ref{tab1} below:

\begin{figure*}[htb]
	\centering
	\includegraphics[width=6.0in]{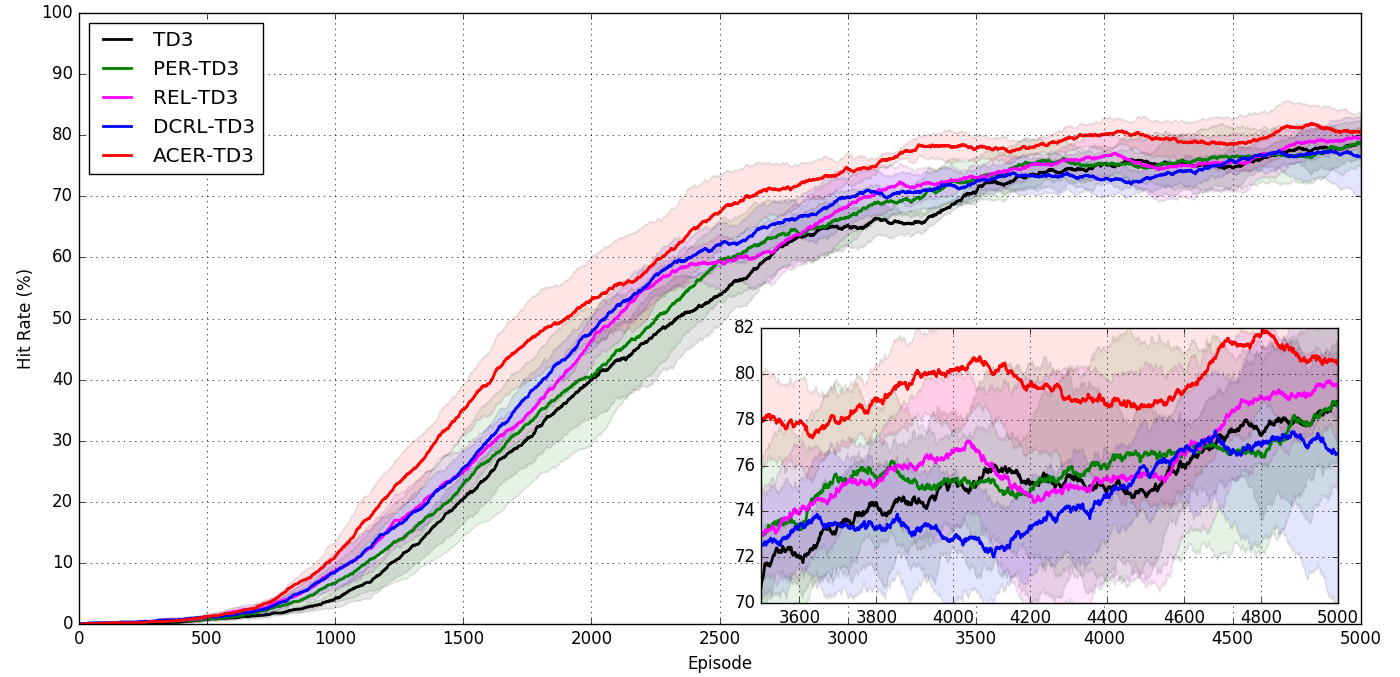}
	\caption{The hit rates of all algorithms. (The subpicture in the lower right corner is an enlarged picture of hit rates in 3,500-5,000 episodes).
	}
	\label{fig13}
\end{figure*}

\begin{table}[htb]	
	\caption{The overall results of all algorithms. (The arrows are attached to point to the better performance and the best results are marked in \textbf{bold}.)}
	\centering
	\begin{tabular}{m{1.7cm}<{\centering}m{0.7cm}<{\centering}m{0.7cm}<{\centering}m{0.7cm}<{\centering}b{0.7cm}<{\centering}}
		
		\hline
		\rule{0pt}{2.7mm}&TP$\uparrow$ &CT$\downarrow$ &SC$\downarrow$ &CR$\uparrow$\\
		\hline
		TD3     & 78.80\%& 3,451& 1.73& 75.31\%\\
		PER-TD3 & 78.80\%& 3,294& 1.17& 75.83\%\\
		REL-TD3 & 79.72\%& 3,067& 1.68& 76.15\%\\
	    DCRL-TD3& 77.52\%& 3,021& 1.71& 74.58\%\\
	    ACER-TD3& \textbf{81.92\%}& \textbf{2,600}& \textbf{1.14}& \textbf{79.52\%}\\
		\hline
	\end{tabular}
	\label{tab1}	
\end{table}

From the experimental results, we can see that the convergence speed of  vanilla TD3 is the slowest due to the uniform experience replay. Other algorithms with different ER mechanisms all have higher convergence speeds. This also further shows the great influence of ER in the DRL algorithms. The proposed ACER algorithm only needs 2,600 episodes to converge, which is an improvement of 24.66\% compared with the TD3 algorithm (which needs 3,451 episodes). However, compared with two of the state-of-the-art ER algorithms, REL and DCRL, the ACER algorithm can still improve by 15.23\% and 13.94\%, respectively. 

For the convergence results, the hit rate of the ACER algorithm is the highest (79.52\%) among all algorithms. In addition, it is worth mentioning that the hit rate of ACER is always the highest throughout the convergence process. Compared with the TD3 algorithm (75.31\%), the convergence result of ACER is improved by 5.59\%. Compared with other advanced ER algorithms, the hit rate of ACER still shows some improvement. An interesting phenomenon is that the convergence speed of DCRL is second only to ACER among all algorithms, but its convergence result is indeed the worst. There are two reasons why DCRL has a fast convergence rate but poor convergence results: 1) The curriculum factor that changes too quickly prevents the agent from fully learning the experience of various difficulties. 2) The adoption of a coverage penalty introduces errors to the priority.

The stability after convergence is another important indicator to evaluate the performance of the algorithms. The smaller the SC is, the better the initial convergence of the algorithm, which means that the update of the network parameters in the later training stage will not have a greater impact on the performance of the algorithm. Among all algorithms, ACER (1.14) has the smallest SC, followed by PER (1.17). The SCs of REL, DCRL, and TD3 are larger, and the values are 1.68, 1.71, and 1.73, respectively. This means that compared to other algorithms, the ACER algorithm is more likely to converge to the optimal solution directly.

\subsection{Testing in Different Environments}
There are two main factors that affect the complexity of the environment: the velocity of obstacles and the number of obstacles. The greater the velocity and the number of obstacles are, the greater the requirement placed on the accuracy or effectiveness of UAV AMC will be. In this section, environments of different levels of complexities are set to explore the generalization capabilities of agents trained by different algorithms. In addition, by drawing the UAV motion trajectories, the performance of the agents in different environments is analyzed and discussed.

\subsubsection{Environments with Different Velocities of Obstacles}
\begin{figure}[htb]
	\centering
	\includegraphics[width=3.5in]{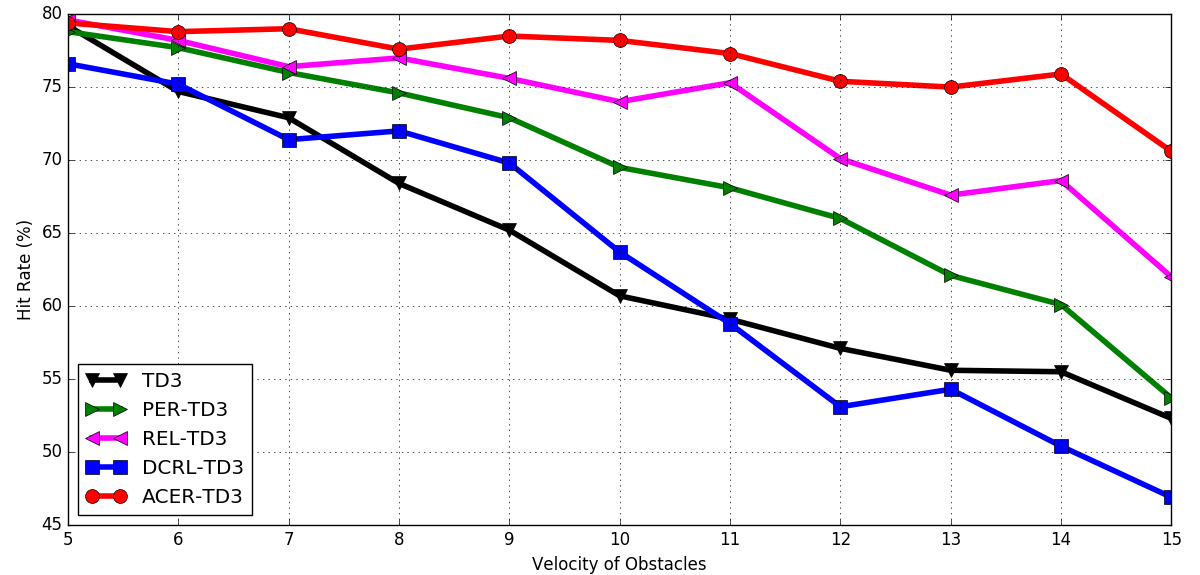}
	\caption{The hit rates of all algorithms with different velocities of obstacles. The velocity of obstacles increases gradually from 5$\mathrm{~m/s}$ to 15$\mathrm{~m/s}$. Each piece of data is calculated by the UAV agent trained by the corresponding algorithm tested in the environment for 1,000 episodes.
	}
	\label{fig14}
\end{figure}

Experiments are conducted in a complex environment with 20 obstacles, and the experimental results are shown in Fig. \ref{fig14}. Obviously, it can be seen that as the speed of obstacles continues to increase, the hit rates of the agents generally show a downward trend. Among these agents, ACER's hit rate dropped the slowest, while DCRL and TD3's hit rates dropped the fastest. When the obstacles’ movement velocity reaches 10$\mathrm{~m/s}$, the hit rate of different agents is significantly different: From high to low, they are ACER (78.20\%), REL (74.10\%), PER (69.50\%), DCRL (63.70\%), and TD3 (60.70\%). When the obstacle's movement velocity reaches 15$\mathrm{~m/s}$, the hit rate gap between the agents is more obvious. ACER still has a hit rate of 70.60\%, while TD3 and DCRL drop to 52.30\% and 46.90\%, respectively.

\begin{figure*}[htb]
	\centering
	\includegraphics[width=7.05in]{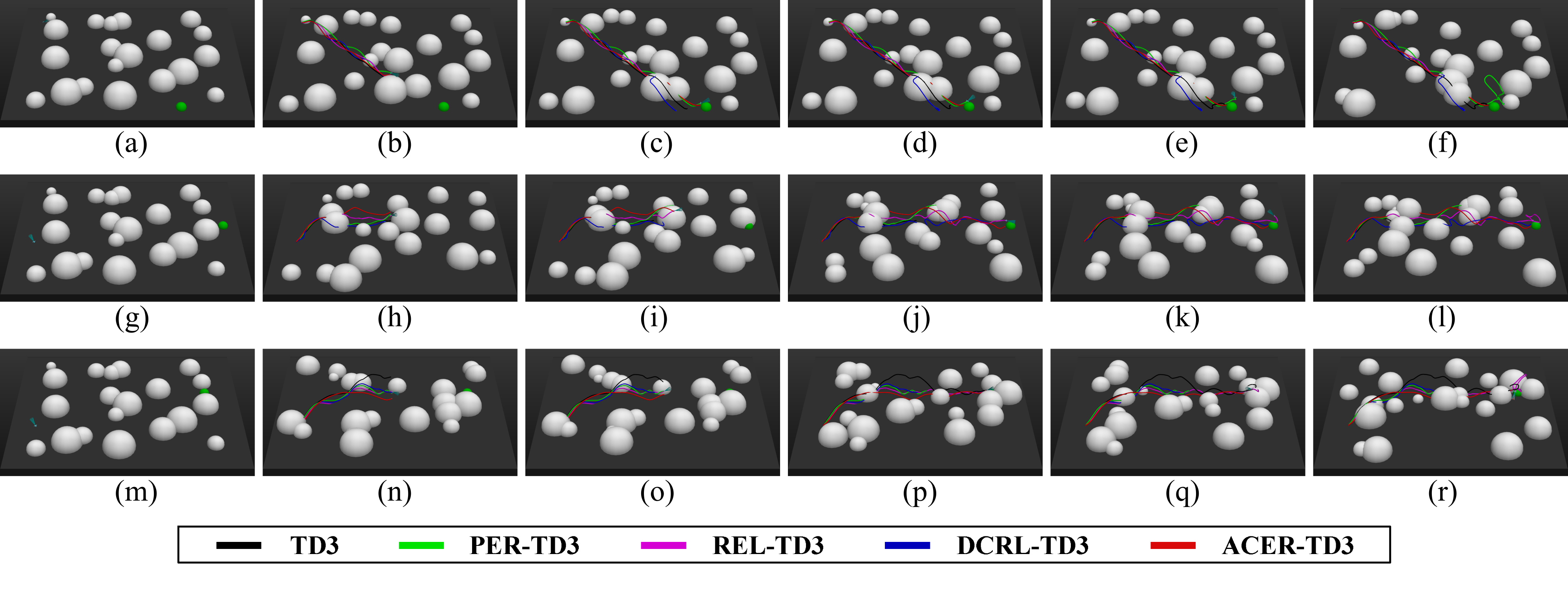}
	\caption{Motion trajectories in environments with different velocities of obstacles. (a). ENV 1. start: (330, 550, 10), end: (-410, -280, 0), velocity of obstacles: 5$\mathrm{~m/s}$. (b). timestep = 948 (REL failed). (c). timestep = 1,352 (ACER succeeded). (d). timestep = 1,384 (DCRL failed). (e). timestep = 1,494 (TD3 succeeded). (f). timestep = 2,082 (PER succeeded). (g). ENV 2. start: (-100, 530, 10), end: (50, -560, 0), velocity of obstacles: 10$\mathrm{~m/s}$. (h). timestep = 916 (TD3 failed). (i). timestep = 1,114 (PER failed). (j). timestep = 1,787 (ACER succeeded). (k). timestep = 1,945 (DCRL succeeded). (l). timestep = 2,183 (REL succeeded). (m). ENV 3. start: (-180, 510, 10), end: (120, -460, 0), velocity of obstacles: 15$\mathrm{~m/s}$. (n). timestep = 910 (DCRL failed). (o). timestep = 1,003 (PER failed). (p). timestep = 1,609 (ACER succeeded). (q). timestep = 1,828 (TD3 failed). (r). timestep = 2,347 (REL succeeded).
	}
	\label{fig15}
\end{figure*}

Fig. \ref{fig15} shows the motion trajectories of all well-trained agents in three environments with different velocities of obstacles. The proposed ACER agent exhibits better generalization ability because it accomplishes the task excellently in all three environments (Fig. \ref{fig15} (c), (j), and (p)). Another agent trained by the advanced REL algorithm performs well in ENV 2 and ENV 3 but takes more timesteps 2,183 (Fig. \ref{fig15} (l)) and 2,347 (Fig. \ref{fig15} (r)), respectively. In test environment ENV 1, two agents (TD3 and PER) also complete the task, which costs 1,494 (Fig. \ref{fig15} (e)) and 2,082 (Fig. \ref{fig15} (f)) timesteps, respectively. However, the DCRL agent is the only other one agent that reaches the target point in ENV 2, taking 1,945 timesteps.

\subsubsection{Environments with Different Numbers of Obstacles}

\begin{figure}[htb]
	\centering
	\includegraphics[width=3.5in]{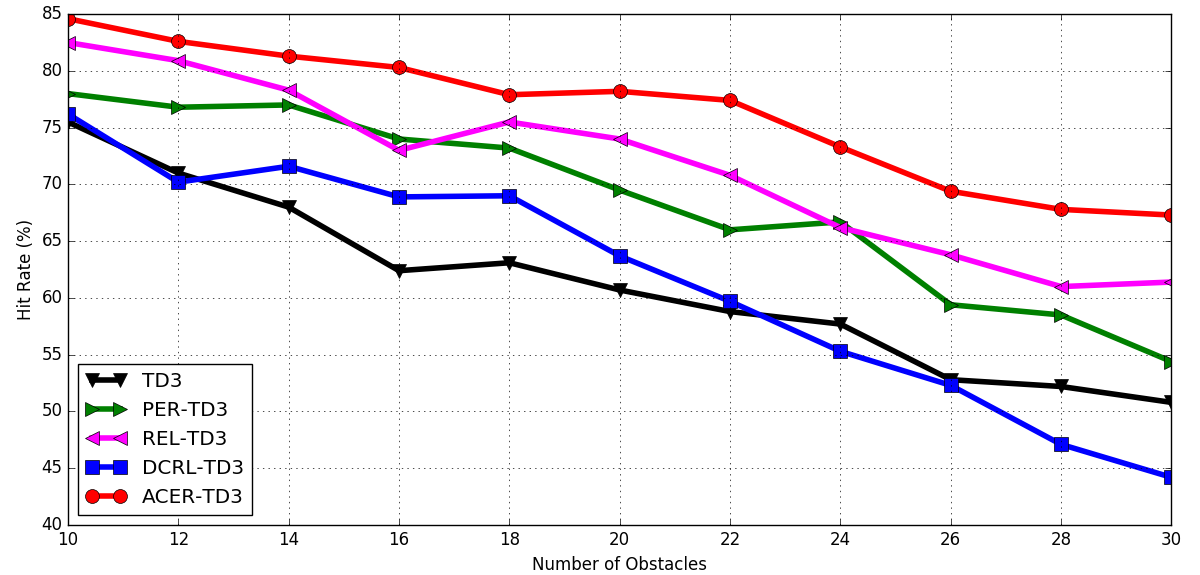}
	\caption{The hit rates of all algorithms with different numbers of obstacles. The number of obstacles increases gradually from 10 to 30. Each piece of data is calculated by the UAV agent trained by the corresponding algorithm tested in the environment for 1,000 episodes.
	}
	\label{fig16}
\end{figure}

The increase in the number of obstacles will lead to a decrease in the density of safe areas in the environment, which will greatly increase the difficulty for the UAV to complete the task. To verify the generalization ability of the algorithms in environments with different obstacle densities, experiments are conducted in an environment in which the velocity of obstacles is 10$\mathrm{~m/s}$. Experimental results (Fig. \ref{fig16}) demonstrate that an increase in the number of obstacles will gradually reduce the hit rates of the agents. As the number of obstacles increased from 10 to 30, the DCRL agent's hit rate dropped the most, from 76.20\% to 44.20\%. The TD3 agent's hit rate dropped from 75.50\% to 50.80\%. Compared with these agents, the ACER agent only has a drop of 17.30\% (from 84.60\% to 67.30\%), which demonstrates excellent generalization ability.

\begin{figure*}[htb]
	\centering
	\includegraphics[width=7.05in]{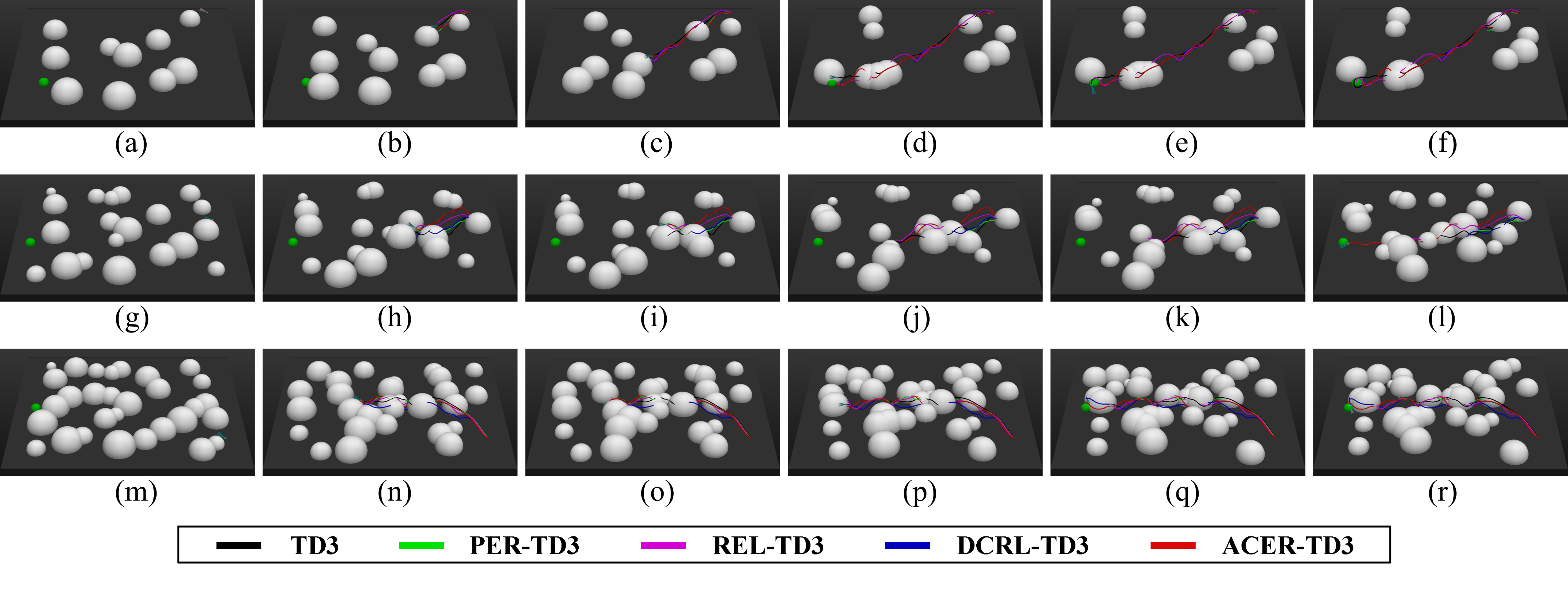}
	\caption{Motion trajectories in environments with different numbers of obstacles. (a). ENV 4. start: (440, -510, 10), end: (-220, 460, 0), number of obstacles: 10. (b). timestep = 386 (PER failed). (c). timestep = 875 (DCRL failed). (d). timestep = 1,580 (ACER succeeded). (e). timestep = 1,676 (REL succeeded). (f). timestep = 1,780 (TD3 succeeded). (g). ENV 5. start: (100, -505, 10), end: (-100, 550, 0), number of obstacles: 20. (h). timestep = 619 (DCRL failed). (i). timestep = 724 (PER failed). (j) timestep = 1,042 (TD3 failed). (k). timestep = 1,143 (REL failed). (l). timestep = 1,730 (ACER succeeded). (m). ENV 6. start: (-280, -530, 10), end: (-20, 530, 0), number of obstacles: 30. (n). timestep = 1,195 (TD3 failed). (o). timestep = 1,315 (PER failed). (p). timestep = 1,571 (REL failed). (q). timestep = 1,840 (ACER succeeded). (r). timestep = 1,905 (DCRL succeeded).
	}
	\label{fig17}
\end{figure*}

Three environments with different numbers of obstacles (Fig. \ref{fig17} (a), (g), and (m)) are set up to show the motion trajectories of the agents trained by all algorithms. In ENV 4 (lowest obstacle density), the ACER, REL, and TD3 agents spend 1,580 (Fig. \ref{fig17} (d)), 1,676 (Fig. \ref{fig17} (e)), and 1,780 (Fig. \ref{fig17} (f)) timesteps flying to the target point and finally succeeding. Only one well-trained agent (ACER) completes the task in ENV 5 (Fig. \ref{fig17} (l)). Except for the ACER agent, which only takes 1,840 timesteps (Fig. \ref{fig17} (q)) to reach the target in the most complex ENV 6, the DCRL agent also completes the task using 1,905 timesteps (Fig. \ref{fig17} (r)). However, the agent trained by the PER algorithm failed in all three environments.

The superiority of the proposed ACER over other algorithms can be demonstrated by the above testing experiments in different environments. The results show the following: 1) Stronger generalization ability: As the complexity of the environment increases, ACER can always maintain the highest hit rate. In addition, the ACER agent is the only agent that successfully completes tasks in six different environments. 2) Higher security: The longer the UAV stays in the enemy area, the more likely it is to be discovered. Compared to other agents, the ACER agent can always take shorter timesteps to complete the task, which ensures that it has higher security. 3) More decisive decision-making ability: An interesting phenomenon that can be easily seen is that TD3 (Fig. \ref{fig17} (f)), PER (Fig. \ref{fig15} (f)), DCRL (Fig. \ref{fig17} (r)), and REL (Fig. \ref{fig15} (l) and (r)) agents sometimes hesitate near the target. This situation usually occurs when the agent detects both the target and the obstacles. The agent wants to approach the target to obtain a larger reward but is afraid of collision with obstacles around the target. This is due to the insufficient learning of the reward function in the training process. The agent is too entangled in the reward of the next few timesteps and does not know that if it flies decisively to the target, it can obtain the reward of success. In a real battlefield environment, UAVs need to make decisive decisions, because once the opportunity is not seized, the UAV may have to wait a long time for the next opportunity. ENV 3 is a good example: Although the REL and TD3 agents also reached the target point early, the target was covered by moving obstacles due to their hesitation. TD3 hovered and finally collided with an obstacle (Fig. \ref{fig15} (q)), while REL took 738 timesteps to wait for the next opportunity (Fig. \ref{fig15} (r)). For UAVs, more decisive decision-making capabilities can not only ensure their own safety, but also increase the success rate of missions. In summary, the excellent performance of the ACER agent shows that the ACER algorithm is a more valuable DRL algorithm when training real UAVs.

\subsection{Additional Exploratory Experiments of Hyperparameters}
The novel ACER algorithm adds some new hyperparameters to the vanilla TD3, and it is necessary to study the influence of the value of these hyperparameters on the performance of the algorithm. In this section, each group of experiments was performed 5 times under different random seeds to calculate the average. We first conducted some groups of experiments in the same training environment with different values of $N_\mathrm{tp}$, and the experimental results are shown in Table \ref{tab2}.
\begin{table}[htb]	
	\caption{Experimental results of different size of temporary pool. (The arrows are attached to point to the better performance and the best results are marked in \textbf{bold}.)}
	\centering
	\begin{tabular}{
			m{0.3cm}<{\centering}
			m{0.4cm}<{\centering}
			m{0.7cm}<{\centering}
			m{0.7cm}<{\centering}
			m{0.7cm}<{\centering}
			m{0.7cm}<{\centering}}
		\hline
		\rule{0pt}{2.7mm}No.&$N_\mathrm{tp}$&TP$\uparrow$&CT$\downarrow$ &SC$\downarrow$ &CR$\uparrow$\\
		\hline
		1&0&78.80\%&3,451&1.73&75.31\%\\
		2&1&78.80\%&3,380&1.98&76.42\%\\
		3&5&79.36\%&\textbf{3,152}&\textbf{1.72}&77.05\%\\
		4&10&\textbf{81.68}\%&3,335&2.04&\textbf{77.79}\%\\
		5&20&76.72\%&3,462&1.90&74.31\%\\
		\hline
	\end{tabular}
	\label{tab2}	
\end{table}

It can be clearly seen that the introduction of the temporary pool can effectively accelerate the convergence of the TD3 algorithm. Different $N_\mathrm{tp}$ has different effects on the results of TD3. If $N_\mathrm{tp}$ is too small (group No. 2), the new experience cannot be fully utilized. If $N_\mathrm{tp}$ is too large (group No. 5), too many continuous experiences will be learned and affect the convergence result. Experiments have found that the values of $N_\mathrm{tp}$ from 5 to 10 can produce better results for the algorithm.

The number of experiences whose priorities are asynchronously updated by the subthread should also be studied. From Table \ref{tab3}, as $A$ increases, the convergence speed and convergence result of the ACER algorithm gradually increase. As $A$ increases past 256, the performance of the algorithm is no longer significantly improved.

\begin{table}[htb]	
	\caption{Experimental results of different asynchronous update numbers. (The arrows are attached to point to the better performance and the best results are marked in \textbf{bold}.)}
	\centering
		\begin{tabular}{
				m{0.3cm}<{\centering}
				m{0.4cm}<{\centering}
				m{0.7cm}<{\centering}
				m{0.7cm}<{\centering}
				m{0.7cm}<{\centering}
				m{0.7cm}<{\centering}}
			\hline
			\rule{0pt}{2.7mm}No.&$A$&TP$\uparrow$&CT$\downarrow$ &SC$\downarrow$ &CR$\uparrow$\\
			\hline
			1&0&76.72\%&3,228&2.12&73.14\%\\
			2&64&77.68\%&3,038&\textbf{1.04}&75.13\%\\
			3&128&77.60\%&2,951&1.50&75.03\%\\
			4&256&81.92\%&2,600&1.13&\textbf{79.52}\%\\
			5&512&\textbf{82.32}\%&\textbf{2,507}&1.95&79.35\%\\
			\hline
	\end{tabular}
	\label{tab3}	
\end{table}

The introduction of CL also adds some hyperparameters, such as the initial value of the curriculum factor $c_\mathrm{init}$, the increment of the curriculum factor $c_\mathrm{incr}$, and constants $k_1$, and $k_2$. From the experimental results shown in Table \ref{tab4}, increasing or decreasing the value of $c_\mathrm{init}$ will have an adverse effect on the performance of the agent, which indicates that the optimal value of $c_\mathrm{init}$ is approximately 10.0. Experiments also show that appropriately reducing $c_\mathrm{incr}$ can achieve better results (group No. 6). In addition, as the value of $k_{1} / k_{2}$ gradually increases, the performance of the agent worsens.

\begin{table}[htb]	
	\caption{Experimental results of different hyperparameters in CER. (Group No. 1 is the control group. The arrows are attached to point to the better performance and the best results are marked in \textbf{bold}.)}
	\centering
	\begin{tabular}{
			m{0.3cm}<{\centering}
			m{0.4cm}<{\centering}
			m{0.4cm}<{\centering}
			m{0.5cm}<{\centering}
			m{0.5cm}<{\centering}
			m{0.7cm}<{\centering}
			m{0.7cm}<{\centering}
			m{0.7cm}<{\centering}
			m{0.7cm}<{\centering}}
		\hline
		\rule{0pt}{2.7mm}No.&$c_\mathrm{init}$&$c_\mathrm{incr}$&$k_1$&$k_2$&TP$\uparrow$&CT$\downarrow$ &SC$\downarrow$ &CR$\uparrow$\\
		\hline
		1&10.0&1.0&0.01&0.005&81.92\%&\textbf{2,600}&1.14&79.52\%\\
		2&10.0&1.0&0.01&0.003&79.26\%&2,877&1.23&77.71\%\\
		3&10.0&1.0&0.01&0.002&77.43\%&3,133&1.15&76.82\%\\
		4&5.0&1.0&0.01&0.005&78.67\%&3,094&1.31&77.33\%\\
		5&15.0&1.0&0.01&0.005&76.13\%&3,242&1.62&75.24\%\\
		6&10.0&0.5&0.01&0.005&\textbf{82.36\%}&2,713&\textbf{1.09}&\textbf{79.66\%}\\
		7&10.0&1.5&0.01&0.005&72.92\%&3,285&2.04&69.43\%\\
		\hline
	\end{tabular}
	\label{tab4}	
\end{table}

\section{Conclusion}
In this work, we design a DRL framework for controlling UAVs in complex unknown dynamic environments. The UAV AMC problem is formulated as an MDP and a novel DRL algorithm, ACER, is proposed to address it. The ACER algorithm uses multithreading to accelerate the update of priorities of the offline experiences. The true priorities are assigned without using the clip function, a temporary experience pool is designed, and an FIUO experience pool is used to ensure that more effective experiences can be learned. In addition, by integrating CL, ACER changes the random training process of DRL into a training process that proceeds from simple to difficult while ensuring the stability of the algorithm. The experimental results demonstrate the success of ACER in comparison to some state-of-the-art DRL algorithms. In addition, the superiority of ACER has also been presented by generalizing the well-trained agent for different large-scale dynamic 3D environments. In future work, we plan to introduce CL into multiagent DRL to achieve more efficient UAV cluster control.


%

\section*{Acknowledgment}
This study was co-supported by the National Natural Science Foundation of China (No. 62003267 and 61573285), the Aeronautical Science Foundation of China (ASFC) (No. 20175553027), and Natural Science Basic Research Plan in Shaanxi Province of China (No. 2020JQ-220).

\ifCLASSOPTIONcaptionsoff
  \newpage
\fi


\bibliographystyle{IEEEtran}
\bibliography{IEEEabrv,Reference}
%



%

\begin{IEEEbiography}[{\includegraphics[width=1in,height=1.25in,clip,keepaspectratio]{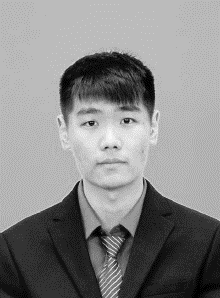}}]{Zijian Hu}
was born in 1996. He received his B.E. in degree detection guidance and control technology from Honors College of Northwestern Polytechnical University (NWPU), Xi'an, China in 2018. He was awarded with an admission from B.E. to Ph.D. directly in 2018. He is currently pursuing the Ph.D. degree in the College of Electronic and Information from NWPU. His current research interests include reinforcement learning theory and the applications of reinforcement learning in UAV control.
\end{IEEEbiography}
\begin{IEEEbiography}[{\includegraphics[width=1in,height=1.25in,clip,keepaspectratio]{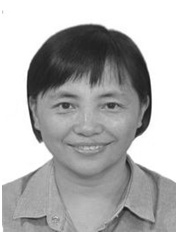}}]{Xiaoguang Gao}
(M'08) was born in 1957. She received her B.E. degree in detection homing and control technology from NWPU in 1982. She completed her master's degree in system engineering from NWPU in 1986. She was awarded the Ph.D. degree from NWPU in 1989. She is currently a professor and the head of the Key Laboratory of Aerospace Information Perception and Photoelectric Control, Ministry of Education, NWPU. Her research interests are machine learning theory, Bayesian network theory, and multiagent control application.
\end{IEEEbiography}
\begin{IEEEbiography}[{\includegraphics[width=1in,height=1.25in,clip,keepaspectratio]{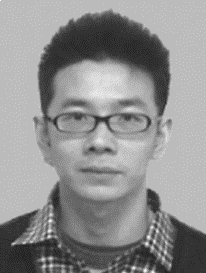}}]{Kaifang Wan}
	was born in 1987. He received his B.E. degree in detection homing and control technology from NWPU, Xi'an, China in 2010. He was awarded with an admission from B.E. to Ph.D. directly in 2010 and he was awarded the Ph.D degree in system engineering in 2016. He is now an assistant researcher of the Key Laboratory of Aerospace Information Perception and Photoelectric Control, Ministry of Education. His current research interests include multiagent theory, approximate dynamic programming and reinforcement learning.
\end{IEEEbiography}
\begin{IEEEbiography}[{\includegraphics[width=1in,height=1.25in,clip,keepaspectratio]{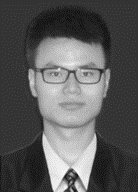}}]{Qianglong Wang}
	 received the B.E. degree in systems engineering from the NWPU, Xi’an, China, in 2017. He is currently working toward the Ph.D degree in control science and engineering at the school of Electronic Information from the NWPU, Xi'an, China. His research interests include deep learning, computer vision and sensitivity analysis.
\end{IEEEbiography}
\begin{IEEEbiography}[{\includegraphics[width=1in,height=1.25in,clip,keepaspectratio]{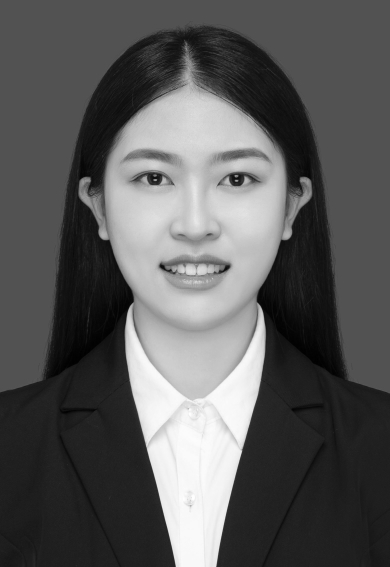}}]{Yiwei Zhai}
	was born in 1997. She received her B.E. degree in 2018 and is currently a postgraduate student at the Department of System Engineering, NWPU, Xi'an, China. Her research interests include path planning, reinforcement learning and multiagent systems.
\end{IEEEbiography}

\vfill


\end{document}